\documentclass[11pt]{article}
\usepackage[margin=1in,footskip=0.25in]{geometry}

\usepackage[utf8]{inputenc}\usepackage[T1]{fontenc}   \usepackage{hyperref}      \usepackage{url}           \usepackage{booktabs}      \usepackage{amsfonts}      \usepackage{nicefrac}      \usepackage{microtype}
\usepackage{subcaption}
\usepackage{amsmath}
\usepackage{bbm}
\usepackage{nccmath}
\usepackage{mathtools}
\usepackage{mathrsfs}
\usepackage{amsbsy}
\usepackage{graphicx}
\usepackage{multirow}
\usepackage{diagbox}
\usepackage{relsize}
\usepackage{xcolor}
\usepackage{amsthm}

\numberwithin{equation}{section}

\newcommand{\wordvec}{{\smaller WORD}2{\smaller VEC}}
\newcommand{\deepwalk}{D{\smaller EEP}W{\smaller ALK}}
\newcommand{\nodevec}{{\smaller NODE}2{\smaller VEC}}
\newcommand{\dyane}{D{\smaller Y}ANE}
\newcommand{\dyngem}{D{\smaller YN}GEM}
\newcommand{\dtriad}{D{\smaller YNAMIC}T{\smaller RIAD}}

\title{Time-varying Graph Representation Learning\\
via Higher-Order Skip-Gram with Negative Sampling}

\author{
  Simone Piaggesi \\
  ISI Foundation, Turin, Italy\\
  University of Bologna, Bologna, Italy\\
  \texttt{simone.piaggesi2@unibo.it} \\
   \and
  Andr\'e Panisson \\
  ISI Foundation, Turin, Italy\\
  \texttt{andre.panisson@isi.it} \\
}

\date{}

\begin{document}

\maketitle

\begin{abstract}
Representation learning models for graphs are
a successful family of techniques that project nodes into feature spaces that can be exploited by other machine learning algorithms.
Since many real-world networks are inherently dynamic, with interactions among nodes changing over time, these techniques can be defined both for static and for time-varying graphs.
Here, we build upon the fact that
the skip-gram embedding approach implicitly performs a matrix factorization, and we extend it to perform implicit tensor factorization on different tensor representations of time-varying graphs.
We show that higher-order skip-gram with negative sampling (HOSGNS) is able to disentangle the role of nodes and time, with a small fraction of the number of parameters needed by other approaches.
We empirically evaluate our approach using time-resolved face-to-face proximity data, showing that the learned time-varying graph representations outperform state-of-the-art methods when used to solve downstream tasks such as network reconstruction, and to predict the outcome of dynamical processes such as disease spreading.
The source code and data are publicly available at \url{https://github.com/simonepiaggesi/hosgns}.
\end{abstract}

\section{Introduction}\label{section-introduction}

A great variety of natural and artificial systems can be represented as networks of elementary structural entities coupled by relations between them.
The abstraction of such systems as networks helps us understand, predict and optimize their behaviour \cite{newman2003structure, albert2002statistical}.
In this sense, node and graph embeddings have been established as standard feature representations in many learning tasks for graphs and complex networks~\cite{cai2018comprehensive, goyal2018graph}.
Node embedding methods map each node of a graph into a low-dimensional vector,
that can be then used to solve downstream tasks
such as edge prediction, network reconstruction and node classification.

Node embeddings have proven successful in achieving low-dimensional encoding of static network structures, but many real-world networks are inherently dynamic, with interactions among nodes changing over time~\cite{holme2012temporal}.
Time-resolved networks are also the support of important dynamical processes, such as epidemic or rumor spreading, cascading failures, consensus formation, etc.~\cite{barrat2008dynamical}
Time-resolved node embeddings have been shown to yield improved performance for
predicting the outcome of dynamical processes over networks, such as information diffusion and disease spreading~\cite{sato2019dyane}.

In this paper we propose a representation learning model that performs an implicit tensor factorization on different higher-order representations of time-varying graphs.
The main contributions of this paper are as follows:

\begin{itemize}
\item  Given that the skip-gram embedding approach implicitly performs a factorization of the shifted pointwise mutual information matrix (PMI)~\cite{levy2014neural}, we generalize it to perform implicit factorization of a shifted PMI tensor. We then define the steps to achieve this factorization using higher-order skip-gram with negative sampling (HOSGNS) optimization.

\item We show how to apply 3rd-order and 4th-order SGNS on different higher-order representations of time-varying graphs.

\item We show that time-varying graph representations learned through HOSGNS outperform state-of-the-art methods when used to solve downstream tasks.

\end{itemize}

We report the results of learning embeddings on empirical time-resolved face-to-face proximity data and using them as predictors for solving two different tasks: network reconstruction and predicting the outcomes of a SIR spreading process over the network. We compare these results with state-of-the art methods for time-varying graph representation learning.

\section{Preliminaries and Related Work}
\label{section-preliminaries-related-works}

\textbf{Skip-gram representation learning.}
The skip-gram model was designed to compute word embeddings in \wordvec~\cite{mikolov2013distributed}, and afterwards extended to graph node embeddings \cite{perozzi2014deepwalk, tang2015line,  grover2016node2vec}. Levy and Goldberg~\cite{levy2014neural} established the relation between skip-gram trained with negative sampling (SGNS) and traditional low-rank approximation methods \cite{kolda2009tensor, anandkumar2014tensor}, showing the equivalence of SGNS optimization to factorizing a shifted \textit{pointwise mutual information} matrix (PMI) \cite{church1990word}.
This equivalence was later retrieved from diverse assumptions~\cite{assylbekov2019context, allen2019vec, melamud2017information, arora2016latent, li2015word}, and exploited to compute closed form
expressions approximated in different graph embedding models \cite{qiu2018network}.
In this work, we refer to the shifted PMI matrix also as $\mathrm{SPMI}_{\kappa} = \mathrm{PMI} -\log\kappa$, where $\kappa$ is the number of negative samples.

\textbf{Random walk based graph embeddings.}
Given an undirected, weighted and connected graph $\mathcal{G = (V,E)}$
with edges $(i,j) \in \mathcal{E}$, nodes $i,j \in \mathcal{V}$ and adjacency matrix $\mathbf{A}$, graph embedding methods are unsupervised models designed to map nodes into dense $d$-dimensional representations ($d \ll |\mathcal{V}|$) encoding structural properties in a vector space \cite{hamilton2017representation}.
A well known family of approaches based on the skip-gram model consists in sampling random walks from the graph and processing node sequences as textual sentences.
In
\deepwalk~\cite{perozzi2014deepwalk} and \nodevec~\cite{grover2016node2vec}, the skip-gram model
is used to obtain node embeddings from co-occurrences in random walk realizations.
Although the original implementation of \deepwalk~uses hierarchical softmax to compute embeddings, we will refer to the SGNS formulation given by \cite{qiu2018network}.
\\
Since SGNS can be interpreted
as a factorization of the word-context PMI matrix~\cite{levy2014neural},
the asymptotic form of the PMI matrix implicitly decomposed in \deepwalk~can be derived~\cite{qiu2018network}.
Given the 1-step transition matrix $\mathbf{P} = \mathbf{D}^{-1}\mathbf{A}$, where $\mathbf{D} = \text{diag}(d_1, \dots, d_{|\mathcal{V}|})$ and $d_i = \sum_{j \in \mathcal{V}}\mathbf{A}_{ij}$ is the (weighted) node degree, the expected PMI for a node-context pair $(i,j)$ occurring in a $T$-sized window is:
\begin{equation}
\label{eq:deepwalk}
    \mathbb{E}[~\mathrm{PMI}^{\textnormal{\deepwalk}}(i,j)~|~T~] = \frac{\frac{1}{2T}\sum_{r=1}^T \left[ p^*(i)(\mathbf{P}^r)_{ij} + p^*(j) (\mathbf{P}^r)_{ji}  \right]}{p^*(i)~p^*(j)}
\end{equation}
where $p^*(i) = \frac{d_i}{\mathrm{vol}(\mathcal{G})}$ is the unique stationary distribution for random walks~\cite{masuda2017random}.
We will use this expression in Section~\ref{subsection-tne} to build PMI tensors from higher-order graph representations.

\textbf{Time-varying graphs and their algebraic representations.}
\label{subsection-repr}
Time-varying graphs  \cite{holme2012temporal} are defined as triples $\mathcal{H = (V,E,T)}$ , i.e. collections of events $(i, j, k) \in \mathcal{E}$, representing undirected pairwise relations among nodes at discrete times ($i,j \in \mathcal{V}$, $k \in \mathcal{T}$).
$\mathcal{H}$ can be seen as a temporal sequence of static adjacency matrices $\{\mathbf{A}^{(k)}\}_{k \in \mathcal{T}}$ such that $\mathbf{A}^{(k)}_{ij} = \omega(i,j,k) \in \mathbb{R}$ is the weight of the event $(i,j,k) \in \mathcal{E}$.
We can concatenate the list of time-stamped snapshots $[\mathbf{A}^{(1)}, \dots, \mathbf{A}^{(|\mathcal{T}|)}]$ to
obtain a single 3rd-order tensor $\boldsymbol{\mathcal{A}}^{stat}(\mathcal{H}) \in \mathbb{R}^{|\mathcal{V}|\times|\mathcal{V}|\times|\mathcal{T}|}$ which characterize the evolution of the graph over time. This representation has been used to discover latent community structures of temporal graphs \cite{gauvin2014detecting} and to perform temporal link prediction
\cite{dunlavy2011temporal}.
Indeed, beyond the above stacked graph representation, more exhaustive representations are possible. In particular, the multi-layer approach \cite{de2013mathematical} allows to map the topology of a time-varying graph $\mathcal{H}$ into a static network $\mathcal{G_{\mathcal{H}}} = (\mathcal{V}_{\mathcal{H}}, \mathcal{E}_{\mathcal{H}})$ (the \textit{supra-adjacency} graph) such that vertices of $\mathcal{G_{\mathcal{H}}}$ correspond to pairs  $(i, k)\equiv i^{(k)} \in \mathcal{V} \times \mathcal{T}$ of the original time-dependent network. This representation can be stored in a 4th-order tensor $\boldsymbol{\mathcal{A}}^{dyn}(\mathcal{H}) \in \mathbb{R}^{|\mathcal{V}|\times|\mathcal{V}|\times|\mathcal{T}|\times|\mathcal{T}|}$ equivalent, up to an opportune reshaping, to the adjacency matrix $\mathbf{A}(\mathcal{G}_{\mathcal{H}}) \in \mathbb{R}^{|\mathcal{V}||\mathcal{T}|\times|\mathcal{V}||\mathcal{T}|}$  associated to $\mathcal{G_{\mathcal{H}}}$. Multi-layer representations for time-varying networks have been used to study time-dependent centrality measures \cite{taylor2019supracentrality} and properties of spreading processes~\cite{valdano2015analytical}.

\textbf{Time-varying graph representation learning.} Given a time-varying graph $\mathcal{H = (V, E, T)}$, we denote as temporal network embedding every model capable to learn from data, implicitly or explicitly, a mapping function:
\begin{equation}
    f : (v, t) \in \mathcal{V} \times \mathcal{T} \mapsto \mathbf{v}^{(t)} \in  \mathbb{R}^d
\label{eq:tempemb}
\end{equation}
which project time-stamped nodes into a latent low-rank vector space that encodes structural and temporal properties of the original evolving graph.\\
Many existing methods learn node representations from sequences of static snapshots through incremental updates in a streaming scenario: deep autoencoders \cite{goyal2018dyngem}, SVD \cite{zhang2018timers}, skip-gram \cite{du2018dynamic} and random walk sampling \cite{beres2019node, mahdavi2018dynnode2vec, yu2018netwalk}.
Another class of models learn dynamic node representations by recurrent/attention mechanisms \cite{goyal2020dyngraph2vec, li2018deep, sankar2020dysat} or by imposing temporal stability among adjacent time intervals \cite{zhou2018dynamic, zhu2016scalable}. \dyane~\cite{sato2019dyane} and {\smaller WEG}2{\smaller VEC}~\cite{torricelli2020weg2vec}
project the dynamic graph structure into a static graph, in order to compute embeddings with \wordvec. Closely related to these ones are \cite{zhan2020si, nguyen2018continuous}, which learn node vectors according to time-respecting random walks or spreading trajectory paths.

The method proposed in \dyane~computes,
given a node $i \in \mathcal{V}$, one vector representation for each time-stamped node $i^{(t)} \in \mathcal{V}^{(\mathcal{T})} = \{(i,t) \in \mathcal{V}\times\mathcal{T}: \exists~ (i,j,t) \in \mathcal{E}\}$ of a supra-adjacency representation $\mathcal{G}_{\mathcal{H}}$ which involves \textit{active nodes} of $\mathcal{H}$.
This representation is inspired by  \cite{valdano2015analytical}, and the supra-adjacency matrix $\mathbf{A}(\mathcal{G}_{\mathcal{H}})$ is defined by two rules:
\begin{enumerate}
    \item For each event $(i,j,t_0)$, if $i$ is also active at time $t_1 > t_0$ and in no other time-stamp between the two, we add a \textit{cross-coupling} edge between supra-adjacency nodes $j^{(t_0)}$ and $i^{(t_1)}$.
    In addition, if the next interaction of $j$ with other nodes happens at $t_2>t_0$, we add an edge between $i^{(t_0)}$ and $j^{(t_2)}$. The weights of such edges are set to $\omega(i,j,t_0)$.
    \item For every case as described above, we also add \textit{self-coupling} edges $(i^{(t_0)}, i^{(t_1)})$ and  $(j^{(t_0)}, j^{(t_2)})$, with weights set to 1.
\end{enumerate}
We will refer to this supra-adjacency representation in Section~\ref{subsection-tne}.
In this representation,
random itineraries correspond to temporal paths of the original time-varying graph, therefore random walk based methods (in particular \deepwalk) are eligible to be used because they give a suitable way to learn node representations according to nodes occurrences observed in such paths.

Some methods learn a single vector representation for each node, squeezing its behaviour over all times, resulting in a quantity $\mathcal{O}(|\mathcal{V}|)$ of embedding parameters.
On the other hand models that learn time-resolved node representations require a quantity $\mathcal{O}(|\mathcal{V}|\times|\mathcal{T}|)$ of embedding parameters
to represent the system in the latent space.
Compared with these methods, our approach requires a quantity $\mathcal{O}(|\mathcal{V}| + |\mathcal{T}|)$ of embedding parameters for disentangled node and time representations.

\section{Proposed Method}
\label{section-method}

Given a time-varying graph $\mathcal{H = (V, E, T)}$,
we propose
a representation learning method that learns disentangled representations for nodes and time slices.
More formally,
we learn a function:
\begin{equation*}
    f^{\ast}:  (v,t) \in \mathcal{V} \times \mathcal{T} \mapsto \mathbf{v}, \mathbf{t} \in  \mathbb{R}^d
\end{equation*}

through a number of parameters proportional to $\mathcal{O}(|\mathcal{V}|+|\mathcal{T}|)$.
This embedding representation can then be reconciled with the definition in Eq.~(\ref{eq:tempemb}) by combining $\mathbf{v}$ and $\mathbf{t}$ in a single $\mathbf{v}^{(t)}$ representation using any combination function $c: (\mathbf{v},\mathbf{t}) \in \mathbb{R}^d \times \mathbb{R}^d \mapsto \mathbf{v}^{(t)} \in  \mathbb{R}^d$.

Starting from the existing skip-gram framework for node embeddings,
we propose a higher-order generalization of skip-gram with negative sampling (HOSGNS) applied to time-varying graphs. We show that this extension allows to implicitly factorize into latent variables higher-order relations that characterize tensor representations of time-varying graphs, in the same way that the classical SGNS decomposes dyadic relations associated to a static graph.
Similar approaches have been applied in NLP for dynamic word embeddings \cite{rudolph2018dynamic}, and higher-order extensions of the skip-gram model have been proposed to learn context-dependent \cite{liu2015learning} and syntactic-aware \cite{cotterell2017explaining} word representations. Moreover tensor factorization techniques have been applied to include the temporal dimension in recommender systems \cite{xiong2010temporal, wu2019neural} and face-to-face contact networks \cite{sapienza2015detecting, gauvin2014detecting}. But this work is the first to merge SGNS with tensor factorization, and then apply it to learn time-varying graph embeddings.

\subsection{Higher-order skip-gram with negative sampling as implicit tensor factorization}
\label{subsection-hosgns}
Here we address the problem of generalizing SGNS to learn embedding representations from higher-order co-occurrences. We analyze here the 3rd-order case, giving the description of the general $n$-order case in the Supplementary Information. Later in this work we will focus 3rd and 4th order representations
since these are the most interesting for time-varying graphs.

We consider a set of training samples
$\mathcal{D} = \{(i, j, k), \;
i \in \mathcal{W}, \; j \in \mathcal{C}, \; k \in \mathcal{T}\}$
obtained by collecting co-occurrences among elements from three sets $\mathcal{W}$, $\mathcal{C}$ and $\mathcal{T}$. Since in SGNS we have pairs of node-context $(i, j)$, this is a direct extension of SGNS to three variables, where $\mathcal{D}$ is constructed e.g. through random walks over a higher-order data structure.
We denote as $\#(i,j,k)$ the number of times the triple $(i,j,k)$ appears in $\mathcal{D}$. Similarly we use $\#i = \sum_{j,k}\#(i,j,k)$,\;
$\#j = \sum_{i,k}\#(i,j,k)$ and
$\#k = \sum_{i,j}\#(i,j,k)$ as the number of times each distinct element occurs in $\mathcal{D}$,
with relative frequencies  $P_{\mathcal{D}}(i,j,k)= \frac{\#(i,j,k)}{|\mathcal{D}|}$,
$P_{\mathcal{D}}(i)= \frac{\#i}{|\mathcal{D}|}$,
$P_{\mathcal{D}}(j)= \frac{\#j}{|\mathcal{D}|}$ and
$P_{\mathcal{D}}(k)= \frac{\#k}{|\mathcal{D}|}$.

Optimization is performed as a binary classification task, where the objective is to discern occurrences actually coming from $\mathcal{D}$ from random occurrences.
We define the likelihood for a single observation  $(i,j,k)$ by applying a sigmoid ($\sigma(x) =  (1+e^{-x})^{-1}$) to the higher-order inner product $[\![\cdot]\!]$ of corresponding $d$-dimensional representations:
\begin{equation}
\label{eq:3rdsgns1}
    P[~(i,j,k) \in \mathcal{D} ~|~\mathbf{w}_i,\mathbf{c}_j,\mathbf{t}_k~] = \sigma\big(~[\![\mathbf{w}_i, \mathbf{c}_j, \mathbf{t}_k]\!]~\big) \equiv \sigma\left(~\sum\nolimits_{r=1}^d
    \mathbf{W}_{ir} \mathbf{C}_{jr} \mathbf{T}_{kr}
    ~\right)
\end{equation}
where embedding vectors $\mathbf{w}_i,\mathbf{c}_j, \mathbf{t}_k \in \mathbb{R}^d$ are respectively rows of $\mathbf{W} \in \mathbb{R}^{|\mathcal{W}| \times d}$, $\mathbf{C} \in \mathbb{R}^{|\mathcal{C}| \times d}$ and $\mathbf{T} \in \mathbb{R}^{|\mathcal{T}| \times d}$. In the 4th-order case we will also have a fourth embedding matrix $\mathbf{S} \in \mathbb{R}^{|\mathcal{S}| \times d}$ related to a fourth set $\mathcal{S}$. For negative sampling we fix an observed $(i,j,k) \in \mathcal{D}$ and independently sample $j_\mathcal{N}$ and $k_\mathcal{N}$ to generate $\kappa$ negative examples $(i,j_\mathcal{N},k_\mathcal{N})$. In this way, for a single occurrence $(i,j,k) \in \mathcal{D}$, the expected contribution to the loss is:
\begin{equation}
\label{eq:3rdsgns2}
    \ell(i,j,k) = \log \sigma \big([\![\mathbf{w}_i, \mathbf{c}_j, \mathbf{t}_k]\!]\big) + \kappa\cdot \underset{j_\mathcal{N}, k_\mathcal{N} \sim P_{\mathcal{N}}}{\mathbb{E}}\Big[\log \sigma \big(-[\![\mathbf{w}_i, \mathbf{c}_{j_\mathcal{N}}, \mathbf{t}_{k_\mathcal{N}}]\!]\big)\Big]
\end{equation}
where the noise distribution is the product of independent marginal probabilities
$P_{\mathcal{N}}(j, k)= P_{\mathcal{D}}(j) \cdot P_{\mathcal{D}}(k)$. Thus the global objective is the sum of all the quantities of Eq.~(\ref{eq:3rdsgns2}) weighted with the corresponding relative frequency  $P_{\mathcal{D}}(i,j,k)$. The full loss function can be expressed as:
\begin{equation}\label{eq:3rdsgns3}
     \mathcal{L} = - \sum_{i=1}^{ |\mathcal{W}|}\sum_{j=1}^{|\mathcal{C}|}\sum_{k=1}^{ |\mathcal{T}|}\Big[ P_{\mathcal{D}}(i,j,k) \log \sigma \big([\![\mathbf{w}_i, \mathbf{c}_j, \mathbf{t}_k]\!]\big) + \kappa\ P_{\mathcal{N}}(i,j,k) \log \sigma \big(-[\![\mathbf{w}_i, \mathbf{c}_j, \mathbf{t}_k]\!]\big)\Big]
\end{equation}
In Supplementary Information we show the steps to obtain Eq.~(\ref{eq:3rdsgns3}) and that it can be optimized with respect to the embedding parameters, satisfying the
low-rank tensor factorization~\cite{kolda2009tensor}
of the multivariate shifted PMI tensor
into factor matrices $\mathbf{W}, \mathbf{C}, \mathbf{T}$:
\begin{equation}
     \sum\nolimits_{r=1}^d
     \mathbf{W}_{ir} \mathbf{C}_{jr} \mathbf{T}_{kr}
     \approx  \log \left(\frac{P_{\mathcal{D}}(i,j,k)}{P_\mathcal{N}(i,j,k)}\right) - \log \kappa \equiv \mathrm{SPMI}_{\kappa}(i,j,k)
\end{equation}

\subsection{Time-varying graph embedding via HOSGNS}
\label{subsection-tne}

\begin{figure}[ht]
    \centering
    \includegraphics[width=.95\linewidth,trim={1.5cm 4.7cm 1.5cm 5.5cm},clip]{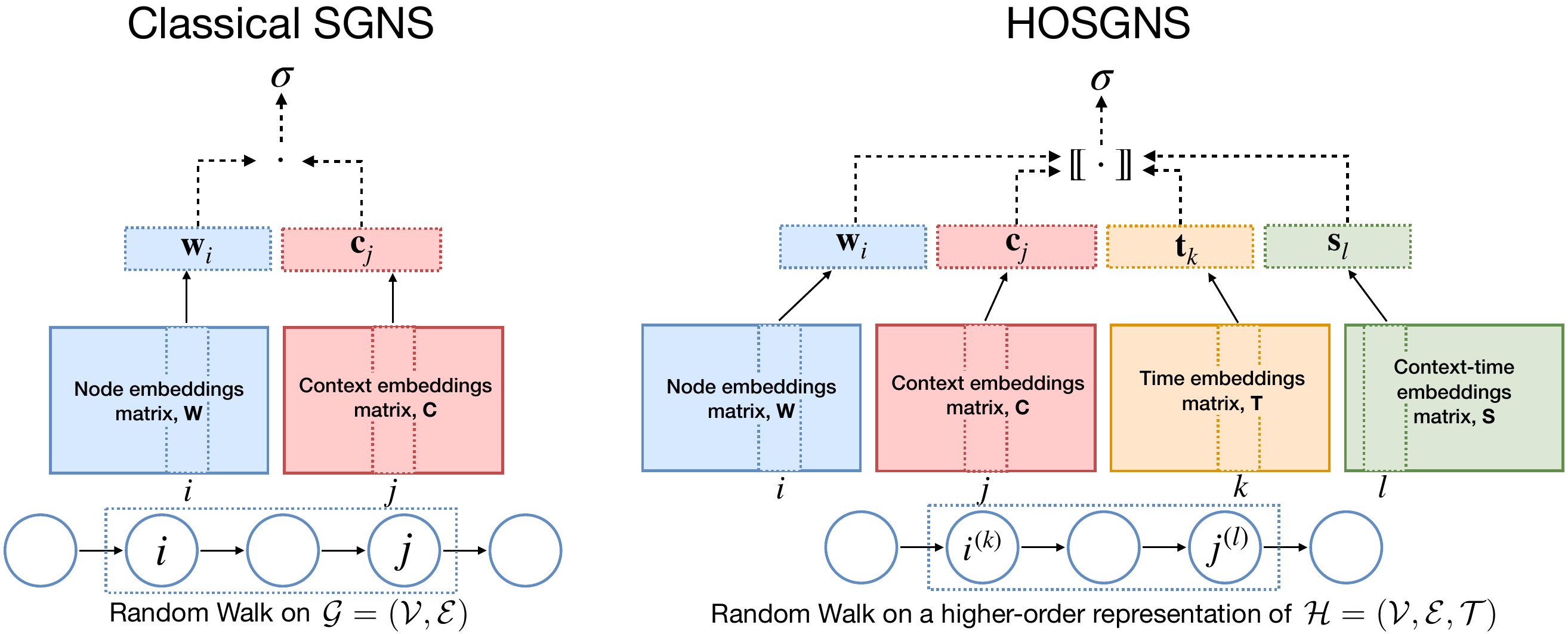}
\caption{Representation of SGNS and HOSGNS with embedding matrices and operations on embedding vectors.
Starting from a random walk realization on a static graph $\mathcal{G=(V,E)}$,
SGNS
takes as input nodes $i$ and $j$ within a context window of size $T$, and maximizes $\sigma(\mathbf{w}_i \cdot \mathbf{c}_j)$. HOSGNS starts from a random walk realization on a  higher-order representation of time-varying graph $\mathcal{H=(V,E,T)}$, takes as input nodes $i^{(k)}$ (node $i$ at time $k$) and $j^{(l)}$ (node $j$ at time $l$) within a context window of size $T$ and maximizes $\sigma([\![\mathbf{w}_i, \mathbf{c}_j, \mathbf{t}_k, \mathbf{s}_l]\!])$. In both cases,
for each input sample, we fix $i$ and draw $\kappa$ combinations of $j$ or $j,k,l$ from a noise distribution,
and we
maximize $\sigma(-\mathbf{w}_i \cdot \mathbf{c}_j)$ (SGNS) or $\sigma(-[\![\mathbf{w}_i, \mathbf{c}_j, \mathbf{t}_k, \mathbf{s}_l]\!])$ (HOSGNS) with their corresponding embedding vectors (negative sampling).
    }
    \label{fig:hosgns}
\end{figure}

While a static graph $\mathcal{G = (V,E)}$ is uniquely represented by an adjacency matrix $\mathbf{A}(\mathcal{G}) \in \mathbb{R}^{|\mathcal{V}|\times|\mathcal{V}|}$, a time-varying graph $\mathcal{H = (V,E,T)}$ admits diverse possible higher-order adjacency relations
(Section \ref{subsection-repr}).
Starting from these higher-order relations, we can either use them directly or use random walk realizations to build a dataset of higher-order co-occurrences.
In the same spirit that random walk realizations give place to co-occurrences that are used to learn embeddings in SGNS, we use higher-order co-occurrences to learn embeddings via HOSGNS.
Figure~\ref{fig:hosgns} summarizes the differences between graph embedding via classical SGNS and time-varying graph embedding via HOSGNS.

As discussed in Section~\ref{subsection-hosgns}, the statistics of higher-order relations can be summarized in the so-called multivariate PMI tensors, which derive from proper co-occurrence probabilities among elements.
Once such PMI tensors are constructed, we can again factorize them via HOSGNS.
To show the versatility of this approach, we choose
PMI tensors derived from two different types of higher-order relations:
\begin{itemize}
    \item A 3rd-order tensor $\boldsymbol{\mathcal{P}}^{(stat)}(\mathcal{H}) \in \mathbb{R}^{|\mathcal{V}|\times|\mathcal{V}|\times|\mathcal{T}|}$
    which gather relative frequencies of nodes occurrences in temporal edges:
    \begin{equation}
    \label{eq:stat-supra}
    (\boldsymbol{\mathcal{P}}^{(stat)})_{ijk} = \frac{\omega(i,j,k)}{\mathrm{vol}(\mathcal{H})}
    \end{equation}
    where $\mathrm{vol}(\mathcal{H}) = \sum_{i,j,k}\omega(i,j,k)$ is the total weight of interactions occurring in $\mathcal{H}$. These probabilities are associated to the snapshot sequence representation $\boldsymbol{\mathcal{A}}^{stat}(\mathcal{H}) = [\mathbf{A}^{(1)}, \dots, \mathbf{A}^{(|\mathcal{T}|)}]$ and contain information about the topological structure of $\mathcal{H}$.

    \item A 4th-order tensor $\boldsymbol{\mathcal{P}}^{(dyn)}(\mathcal{H}) \in \mathbb{R}^{|\mathcal{V}|\times|\mathcal{V}|\times|\mathcal{T}|\times|\mathcal{T}|}$, which gather occurrence probabilities of time-stamped nodes over random walks
    of the supra-adjacency graph $\mathcal{G}_{\mathcal{H}}$ proposed in \cite{valdano2015analytical} (as in \dyane). Using the numerator of Eq.~(\ref{eq:deepwalk}) tensor entries are given by:
    \begin{equation}
    \label{eq:dyn-supra}
       (\boldsymbol{\mathcal{P}}^{(dyn)})_{ijkl} = \frac{1}{2T}\sum_{r=1}^T \left[ \frac{d_{(ik)}}{\mathrm{vol}(\mathcal{G}_{\mathcal{H}})}(\mathbf{P}^r)_{(ik)(jl)} + \frac{d_{(jl)}}{\mathrm{vol}(\mathcal{G}_{\mathcal{H}})} (\mathbf{P}^r)_{(jl)(ik)} \right]
    \end{equation}
    where $(ik)$ and $(jl)$ are lexicographic indices of the supra-adjacency matrix $\mathbf{A}(\mathcal{G}_{\mathcal{H}})$ corresponding to nodes $i^{(k)}$ and node $j^{(l)}$. These probabilities encode causal dependencies among temporal nodes and are correlated with dynamical properties    of spreading processes.
\end{itemize}

We also combined the two representations
in a single tensor that is the average of $\boldsymbol{\mathcal{P}}^{(stat)}$ and $\boldsymbol{\mathcal{P}}^{(dyn)}$
\begin{equation}
    (\boldsymbol{\mathcal{P}}^{(stat|dyn)})_{ijkl} = \frac{1}{2}\left[(\boldsymbol{\mathcal{P}}^{(stat)})_{ijk} \delta_{kl} + (\boldsymbol{\mathcal{P}}^{(dyn)})_{ijkl}\right]
    \label{eq:statdyn-supra}
\end{equation}
where $\delta_{kl} = \mathbbm{1}[k=l]$ is the Kronecker delta.
In this framework indices $(i,j,k)$  correspond to triples \textit{(node, context, time)} and indices $(i,j,k,l)$ correspond to \textit{(node, context, time, context-time)}.

The above tensors
gather empirical probabilities $P_\mathcal{D}(i,j,k\dots)$ corresponding to positive examples of observable higher-order relations.
The probabilities of negative examples $P_\mathcal{N}(i,j,k\dots)$ can be obtained as the product of marginal distributions $P_\mathcal{D}(i), P_\mathcal{D}(j), P_\mathcal{D}(k)\dots$~Computing exactly the objective function in Eq.~(\ref{eq:3rdsgns3}) (or the 4th-order analogous) is computationally expensive, but it can be approximated by a sampling strategy:
picking positive tuples according to the data distribution $P_\mathcal{D}$ and negative ones according to independent sampling $P_\mathcal{N}$, HOSGNS objective can be asymptotically approximated through the optimization of the following weighted loss:
\begin{equation}
    \label{eq:cross-ent}
    -\frac{1}{B}\Big[\sum_{(ijk\dots) \sim P_\mathcal{D}}^B \log \sigma \big([\![\mathbf{w}_i, \mathbf{c}_j, \mathbf{t}_k,\dots]\!]\big) + \kappa\cdot\sum_{(ijk\dots) \sim P_\mathcal{N}}^B \log \sigma \big(-[\![\mathbf{w}_i, \mathbf{c}_j, \mathbf{t}_k, \dots]\!]\big) \Big]
\end{equation}
where $B$ is the number of the samples drawn in a training step and $\kappa$ is the negative sampling constant.

\section{Experiments}\label{section-experiments}

For our experiments we use time-varying graphs
collected by the SocioPatterns collaboration (http://www.sociopatterns.org) using wearable proximity sensors that sense the face-to-face proximity relations of individuals wearing them.
After training the proposed models (HOSGNS applied to $\boldsymbol{\mathcal{P}}^{(stat)}$ , $\boldsymbol{\mathcal{P}}^{(dyn)}$ or $\boldsymbol{\mathcal{P}}^{(stat|dyn)}$) on each dataset,
we extract from embedding matrices $\mathbf{W, C, T, S}$ (the latter not in the case of $\boldsymbol{\mathcal{P}}^{(stat)}$) the embedding vectors
$\mathbf{w}_i, \mathbf{c}_j, \mathbf{t}_k, \mathbf{s}_l$
where $i,j \in \mathcal{V}$ and $k,l \in \mathcal{T}$
and we use them to solve different downstream tasks: \textit{node classification} and \textit{temporal event reconstruction}.

\subsection{Experimental Setup}

\newcommand{\LyonSchool}{{L}{\smaller YON}{S}{\smaller CHOOL}}
\newcommand{\InVS}{{I}{\smaller N}{VS15}}
\newcommand{\Thiers}{T{\smaller HIERS}13}

\textbf{Datasets.} We used publicly available data sets describing face-to-face proximity of individuals with a temporal resolution of 20 seconds~\cite{cattuto2010dynamics}. These datasets were collected by the SocioPatterns collaboration in a variety of contexts, namely in a school (``\LyonSchool''),  a conference (``SFHH''), a hospital (``LH10''), a highschool (``\Thiers''), and in offices (``\InVS'')~\cite{genois2018can}.
To our knowledge, this is the largest collection of open data sets sensing proximity in the same range and temporal resolution that are being used by modern contact tracing systems.
We built a time-varying graph from each dataset by aggregating the data on 600 seconds time windows, and neglecting those snapshots without registered interactions at that time scale. If multiple events are recorded between nodes $(i,j)$ in a certain aggregated window $k$, we denote the weight of the link $(i,j,k)$ with the number of such interactions. Table~\ref{tab:datastats} shows some basic statistics for each data set.

\begin{table}[ht!]
\caption{Summary statistics about empirical time-varying graph data. In order: number of single nodes $|\mathcal{V}|$, number of steps $|\mathcal{T}|$, number of events $|\mathcal{E}|$, number of active nodes $|\mathcal{V}^{(\mathcal{T})}|$, average weight of events $\frac{1}{|\mathcal{E}|}\sum_{e \in \mathcal{E}} \omega(e)$, nodes density  $\frac{|\mathcal{V}^{(\mathcal{T})}|}{|\mathcal{V}||\mathcal{T}|}$ and links density $\frac{2|\mathcal{E}|}{|\mathcal{V}|(|\mathcal{V}|-1)|\mathcal{T}|}$.}
    \makebox[\linewidth]{
    \centering
    \begin{small}
    \begin{tabular}{lccccccc}
         \toprule
         Dataset & $|\mathcal{V}|$ &  $|\mathcal{T}|$ & $|\mathcal{E}|$ & $|\mathcal{V}^{(\mathcal{T})}|$ & Average Weight & Nodes Density & Links Density  \\
         \midrule
         \LyonSchool & 242 & 104 & 44820 & 17174  & 2.806 & 0.6824 & 0.0148 \\
         SFHH & 403 & 127 & 17223 & 10815  & 4.079 & 0.2113 & 0.0017 \\
         LH10 & 76 & 321 & 7435 & 4880  & 4.448 & 0.2000 & 0.0081 \\
         \Thiers & 327 & 246 & 35862 & 32546  & 5.256 & 0.4046 & 0.0027 \\
         \InVS & 217 & 691 & 18791 & 22451  & 4.164 & 0.1497 & 0.0012 \\
         \bottomrule
    \end{tabular}
    \end{small}
    }
    \label{tab:datastats}
\end{table}

\textbf{Baselines.} We compare our approach with several baseline methods from the literature of time-varying graph embeddings, which learn time-stamped node representations:
\begin{itemize}
    \item \dyane~\textnormal{\cite{sato2019dyane}}. Learns temporal node embeddings with \deepwalk, mapping a time-varying graph into a supra-adjacency representation. As in the original paper, we used the implementation of \nodevec\footnote{\texttt{https://github.com/snap-stanford/snap/tree/master/examples/node2vec}} with $p=q=1$.
    \item \dyngem~\textnormal{\cite{goyal2018dyngem}}. Deep autoencoder architecture which dinamically reconstructs each graph snapshot initializing model weights with parameters learned in previous time frames. We used the code made available online from the authors\footnote{\texttt{http://www-scf.usc.edu/\string~nkamra/}}.
    \item \dtriad~\textnormal{\cite{zhou2018dynamic}}. Captures structural information and temporal patterns of nodes, modeling the \textit{triadic closure} process. We used the reference implementation available in the official repository\footnote{\texttt{https://github.com/luckiezhou/DynamicTriad}}.
\end{itemize}

Details about hyper-parameters used in each method can be found in the Supplementary Information.

\subsection{Downstream tasks}

\textbf{Node Classification.} In this task, we aim to classify nodes in epidemic states according to a SIR epidemic process~\cite{barrat2008dynamical} with
infection rate $\beta$ and recovery rate $\mu$.
We simulated 5 realizations of the SIR process on top of each empirical graph with different combinations of parameters $(\beta,\mu)$.
We used the same combinations of epidemic parameters and the same dynamical process to produce SIR states as described in \cite{sato2019dyane}.
Then we set a logistic regression task to classify epidemic states S-I-R assigned to each active node $i^{(k)}$ during the unfolding of the spreading process.
We combine the embedding vectors of HOSGNS as follows:
for HOSGNS$^{(stat)}$,
we use the Hadamard (element-wise) product $\mathbf{w}_i\circ\mathbf{c}_i\circ\mathbf{t}_k$; for HOSGNS$^{(dyn)}$ and HOSGNS$^{(stat|dyn)}$, we use $\mathbf{w}_i\circ\mathbf{t}_k$.
We compared with dynamic node embeddings learned from baselines. For fair comparison, all models are required produce time-stamped node representations with dimension $d = 128$ as input to the logistic regression.

\textbf{Temporal Event Reconstruction.}
In this task, we aim to determine if an event $(i,j,k)$ is in $\mathcal{H=(V,E,T)}$, i.e., if there is an edge between nodes $i$ and $j$ at time $k$.
We create a random time-varying graph $\mathcal{H^*=(V,E^*,T)}$ with same active nodes $\mathcal{V}^{(\mathcal{T})}$ and a number of $|\mathcal{E}|$ events that are not part of $\mathcal{E}$.
Embedding representations learned from $\mathcal{H}$ are used as features to train
a logistic regression to predict if a given event $(i,j,k)$ is in $\mathcal{E}$ or in $\mathcal{E^*}$.
We combine the embedding vectors of HOSGNS as follows:
for HOSGNS$^{(stat)}$,
we use the Hadamard product $\mathbf{w}_i\circ\mathbf{c}_j\circ\mathbf{t}_k$; for HOSGNS$^{(dyn)}$ and HOSGNS$^{(stat|dyn)}$, we use $\mathbf{w}_i\circ\mathbf{c}_j\circ\mathbf{t}_k\circ\mathbf{s}_k$.
For baseline methods, we aggregate vector embeddings to obtain link-level representations with binary operators (\textit{Average}, \textit{Hadamard}, \textit{Weighted-L1}, \textit{Weighted-L2} and \textit{Concat}) as already used in previous works \cite{grover2016node2vec, tsitsulin2018verse}.
For fair comparison, all models are required produce event representations with dimension $d = 192$ as input to the logistic regression.

Tasks were evaluated using train-test split. To avoid information leakage from training to test,
we randomly split $\mathcal{V}$ and $\mathcal{T}$ in train and test sets $(\mathcal{V}_{tr}, \mathcal{V}_{ts})$ and $(\mathcal{T}_{tr}, \mathcal{T}_{ts})$, with proportion $70\%-30\%$. For node classification, only nodes in $\mathcal{V}_{tr}$ at times in $\mathcal{T}_{tr}$ were included in the train set, and only nodes in $\mathcal{V}_{ts}$ at times in $\mathcal{T}_{ts}$ were included in the test set. For temporal event reconstruction, only events with $i,j \in \mathcal{V}_{tr}$ and $k \in \mathcal{T}_{tr}$ were included in the train set, and only events with $i,j \in \mathcal{V}_{ts}$ and $k \in \mathcal{T}_{ts}$ were included in the test set.

\subsection{Results}

All approaches were evaluated for both downstream tasks in terms of Macro-F1 scores in all datasets.
5 different runs of the embedding model are evaluated on 10 different train-test splits for both downstream tasks. We collect the average with standard deviation over each run of the embedding model, and report the average with standard deviation over all runs.
In node classification, every SIR realization is assigned to a single embedding run to compute prediction scores.

Results for the classification of nodes in epidemic states are shown in Table~\ref{tab:classification}, and are in line with the results reported in~\cite{sato2019dyane}. We report here a subset of $(\beta,\mu)$ but other combinations are available on the Supplementary Information, and they confirm the conclusions discussed here.  \dyngem~and \dtriad~have low scores, since they are not devised to learn from graph dynamics. HOSGNS$^{(stat)}$ is not able to capture the graph dynamics due to the static nature of $\boldsymbol{\mathcal{P}}^{(stat)}$. \dyane, HOSGNS$^{(stat|dyn)}$ and HOSGNS$^{(dyn)}$ show good performance in this task, with these two HOSGNS variants outperforming \dyane~in most of the combinations of datasets and SIR parameters.

\begin{table}[ht!]
\centering
\caption{Macro-F1 scores for classification of nodes in epidemic states according to a SIR model with parameters $(\beta,\mu)$.
For each $(\beta,\mu)$ we highlight the two highest  scores and underline the best one.}
\makebox[\linewidth]{
\begin{scriptsize}
\begin{tabular}{clccccc}\toprule
\multirow{2}{*}{$(\beta,\mu)$}&\multirow{2}{*}{Model}&\multicolumn{5}{c}{Dataset}\\
&&\LyonSchool&SFHH&LH10&\Thiers&\InVS\\

\midrule\multirow{6}{*}{$(0.25,0.002)$}
&\dyane &$77.8\pm1.4$&$66.7\pm2.0$&$54.7\pm2.4$&$\mathbf{73.2\pm1.2}$&$\mathbf{\underline{64.9}\pm1.1}$\\
&\dyngem&$57.3\pm1.5$&$39.9\pm2.3$&$34.7\pm1.9$&$36.8\pm1.5$&$59.0\pm2.3$\\
&\dtriad&$30.9\pm0.7$&$29.1\pm1.0$&$30.3\pm0.8$&$30.5\pm0.3$&$30.6\pm0.3$\\
\cline{2-7}

&HOSGNS$^{(stat)}$&$60.1\pm2.1$&$55.8\pm1.5$&$50.0\pm2.1$&$49.9\pm1.8$&$46.4\pm1.0$\\

&HOSGNS$^{(dyn)}$&$\mathbf{\underline{78.9}\pm1.1}$&$\mathbf{\underline{69.1}\pm1.4}$&$\mathbf{\underline{61.7}\pm1.7}$&$\mathbf{\underline{73.4}\pm1.2}$&$\mathbf{64.4\pm1.4}$\\

&HOSGNS$^{(stat|dyn)}$&$\mathbf{78.6\pm1.1}$&$\mathbf{68.2\pm1.3}$&$\mathbf{61.6\pm2.3}$&$72.2\pm1.3$&$63.8\pm1.4$\\

\midrule\multirow{6}{*}{$(0.125,0.001)$}
&\dyane  &$74.4\pm1.2$&$68.3\pm1.3$&$64.3\pm1.8$&$72.5\pm0.6$&$65.9\pm1.3$\\
&\dyngem &$56.8\pm1.7$&$30.6\pm2.0$&$39.6\pm1.8$&$33.9\pm0.9$&$59.3\pm1.5$\\
&\dtriad &$32.8\pm1.1$&$31.6\pm1.2$&$30.4\pm0.9$&$27.6\pm0.8$&$29.6\pm0.2$\\
\cline{2-7}

&HOSGNS$^{(stat)}$&$60.4\pm1.7$&$55.7\pm1.6$&$50.4\pm1.9$&$54.4\pm0.9$&$47.8\pm1.2$\\

&HOSGNS$^{(dyn)}$&$\mathbf{\underline{76.0}\pm0.8}$&$\mathbf{68.5\pm1.6}$&$\mathbf{65.3\pm2.8}$&$\mathbf{\underline{75.6}\pm0.7}$&$\mathbf{66.8\pm1.3}$\\

&HOSGNS$^{(stat|dyn)}$&$\mathbf{75.1\pm1.3}$&$\mathbf{\underline{68.9}\pm1.3}$&$\mathbf{\underline{66.1}\pm1.8}$&$\mathbf{75.0\pm0.7}$&$\mathbf{\underline{66.9}\pm1.2}$\\
\midrule\multirow{6}{*}{$(0.0625,0.002)$}&\dyane&$73.0\pm1.0$&$64.0\pm1.2$&$53.0\pm2.2$&$66.5\pm0.8$&$\mathbf{\underline{59.8}\pm0.9}$\\
&\dyngem&$54.3\pm1.9$&$32.0\pm1.4$&$33.0\pm1.5$&$33.7\pm0.9$&$53.8\pm1.1$\\
&\dtriad&$29.4\pm0.9$&$30.1\pm1.1$&$30.4\pm0.9$&$27.2\pm0.6$&$28.7\pm0.5$\\
\cline{2-7}

&HOSGNS$^{(stat)}$&$58.5\pm1.8$&$51.6\pm1.2$&$46.0\pm1.5$&$49.4\pm0.8$&$46.5\pm0.8$\\

&HOSGNS$^{(dyn)}$&$\mathbf{\underline{74.4}\pm1.0}$&$\mathbf{\underline{65.1}\pm1.2}$&$\mathbf{56.8\pm1.8}$&$\mathbf{\underline{68.4}\pm0.7}$&$\mathbf{59.6\pm0.9}$\\

&HOSGNS$^{(stat|dyn)}$&$\mathbf{73.1\pm1.2}$&$\mathbf{64.6\pm1.3}$&$\mathbf{\underline{56.9}\pm1.9}$&$\mathbf{67.9\pm0.7}$&$59.4\pm1.0$\\

\bottomrule
\end{tabular}

\end{scriptsize}
}
\label{tab:classification}
\end{table}

Results for the temporal event reconstruction task are reported in Table~\ref{tab:reconstruction}.
Temporal event reconstruction is not performed well by \dyngem. \dtriad~has better performance with Weighted-L1 and Weighted-L2 operators, while \dyane~has better performance using Hadamard or Weighted-L2.
Since Hadamard product is explicitly used in Eq.~(\ref{eq:3rdsgns1}) to optimize HOSGNS, all HOSGNS variants show best scores with this operator.
HOSGNS$^{(stat)}$ outperforms all approaches,
setting new state-of-the-art results in this task. The $\boldsymbol{\mathcal{P}}^{(dyn)}$ representation used as input to HOSGNS$^{(dyn)}$ does not focus on events but on dynamics, so the performance for event reconstruction is slightly below \dyane, while HOSGNS$^{(stat|dyn)}$ is comparable to \dyane. Results for HOSGNS models using other operators are available in the Supplementary Information.

\begin{table}[ht!]
\centering
\caption{Macro-F1 scores for temporal event reconstruction.
We highlight in bold the best two overall scores for each dataset.
For baseline models we underline their highest score.
}
\makebox[\linewidth]{
\begin{scriptsize}
\begin{tabular}{llccccc}\toprule
\multirow{2}{*}{ Model}&\multirow{2}{*}{Operator}&\multicolumn{5}{c}{Dataset}\\
&&\LyonSchool&SFHH&LH10&\Thiers&\InVS\\
\midrule

\multirow{5}{*}{\dyane}
&Average&$56.6\pm0.9$&$52.7\pm1.2$&$53.2\pm1.6$&$51.2\pm0.8$&$52.3\pm1.0$\\
&Hadamard&$89.5\pm0.6$&$\underline{86.5}\pm1.2$&$\underline{73.9}\pm1.5$&$94.4\pm0.3$&$93.7\pm0.4$\\
&Weighted-L1&$89.8\pm0.5$&$83.2\pm1.1$&$72.1\pm1.4$&$95.1\pm0.3$&$94.5\pm0.4$\\
&Weighted-L2&$\underline{90.5}\pm0.6$&$84.2\pm1.0$&$72.5\pm1.4$&$\mathbf{\underline{95.2}}\pm0.2$&$\mathbf{\underline{94.7}\pm0.4}$\\
&Concat&$65.8\pm1.0$&$53.3\pm1.0$&$55.8\pm1.2$&$57.4\pm1.3$&$50.8\pm1.0$\\
\cline{2-7}

\multirow{5}{*}{\dyngem}
&Average&$57.8\pm0.8$&$56.9\pm1.1$&$\underline{54.1}\pm1.8$&$40.1\pm0.6$&$43.2\pm1.4$\\
&Hadamard&$\underline{62.1}\pm0.9$&$54.4\pm1.4$&$52.0\pm2.2$&$39.7\pm1.0$&$44.5\pm1.3$\\
&Weighted-L1&$58.6\pm0.6$&$52.7\pm1.2$&$49.9\pm1.8$&$\underline{41.5}\pm0.5$&$\underline{45.9}\pm1.1$\\
&Weighted-L2&$54.3\pm0.8$&$47.0\pm1.4$&$46.5\pm1.9$&$39.5\pm0.5$&$42.6\pm1.5$\\
&Concat&$60.4\pm0.7$&$\underline{58.2}\pm0.9$&$48.2\pm1.8$&$36.9\pm0.5$&$45.2\pm1.1$\\
\cline{2-7}

\multirow{5}{*}{\dtriad}&Average&$51.4\pm0.6$&$57.0\pm0.9$&$58.4\pm1.4$&$57.7\pm0.5$&$55.1\pm0.7$\\
&Hadamard&$60.9\pm0.5$&$58.7\pm0.8$&$58.6\pm1.3$&$62.2\pm0.4$&$64.3\pm0.7$\\
&Weighted-L1&$\underline{78.7}\pm1.0$&$72.4\pm0.8$&$75.5\pm1.1$&$70.7\pm0.7$&$78.3\pm0.6$\\
&Weighted-L2&$77.1\pm0.9$&$\underline{72.9}\pm1.3$&$\mathbf{\underline{77.0}}\pm1.1$&$\underline{72.3}\pm0.6$&$\underline{78.7}\pm0.7$\\
&Concat&$52.5\pm0.6$&$53.3\pm0.9$&$56.3\pm1.1$&$55.1\pm0.5$&$52.9\pm0.8$\\
\midrule

HOSGNS$^{(stat)}$
&Hadamard&$\mathbf{{99.4}\pm0.1}$&$\mathbf{{98.5}\pm0.2}$&$\mathbf{{99.6}\pm0.2}$&$\mathbf{{99.4}\pm0.1}$&$\mathbf{{98.6}\pm0.3}$\\

HOSGNS$^{(dyn)}$
&Hadamard&$89.8\pm0.6$&$81.9\pm0.9$&$70.0\pm1.3$&$92.6\pm0.3$&$86.4\pm0.6$\\

HOSGNS$^{(stat|dyn)}$
&Hadamard&$\mathbf{92.6\pm0.3}$&$\mathbf{87.5\pm0.8}$&$76.9\pm1.5$&$94.2\pm0.3$&$89.3\pm0.5$\\
\bottomrule
\end{tabular}

\end{scriptsize}

}
\label{tab:reconstruction}
\end{table}

We observe an overall good performance of HOSGNS$^{(stat|dyn)}$ in both downstream tasks, being in almost all cases the second highest score, compared to the other two variants which excel in one task but fail in the other one.
One of the main advantages of HOSGNS is that this methodology is able to disentangle the role of nodes and time by learning representations of nodes and time intervals separately. While models that learn node-time representations (such as \dyane) need a number of parameters that is at least $\mathcal{O}(|\mathcal{V}|\times|\mathcal{T}|)$, HOSGNS is able to learn node and time representations separately, with a number of parameters in the order of $\mathcal{O}(|\mathcal{V}| + |\mathcal{T}|)$.
In the Supplementary Information we include plots with two dimensional projections of these embeddings, showing that the embedding matrices of HOSGNS approaches successfully capture both the structure and the dynamics of the time-varying graph.

\section{Conclusions}\label{section-conclusions}

In this paper, we introduce higher-order skip-gram with negative sampling (HOSGNS) for time-varying graph representation learning.
We show that this method is able to disentangle the role of nodes and time, with a small fraction of the number of parameters needed by other methods.
The embedding representations learned by HOSGNS outperform other methods in the literature and set new state-of-the-art results for predicting the outcome of dynamical processes and for temporal event reconstruction.
We show that HOSGNS can be intuitively applied to time-varying graphs, but this methodology can be easily adapted to solve other representation learning problems that involve  multi-modal data and multi-layered graph representations.

\section*{Acknowledgments}

The authors would like to thank Prof. Ciro Cattuto for the fruitful discussions that helped shaping this manuscript.
AP  acknowledges  partial  support  from  Research  Project  Casa  Nel  Parco  (POR
FESR 14/20 - CANP - Cod.  320 - 16 - Piattaforma Tecnologica Salute e Benessere) funded
by Regione Piemonte in the context of the Regional Platform on Health and Wellbeing and
from  Intesa  Sanpaolo  Innovation  Center.   The  funders  had  no  role  in  study  design,  data
collection and analysis, decision to publish, or preparation of the manuscript.

\bibliographystyle{ieeetr}
{\small\bibliography{arxiv}}
\newpage
\appendix

\section{Supplementary Information}

\subsection{Low-rank tensor decomposition}

Low-rank tensor decomposition \cite{kolda2009tensor} aims to factorize a generic  tensor into a sum of rank-one tensors. For example, given a 3rd-order tensor $\boldsymbol{\mathcal{X}} \in \mathbb{R}^{I \times J \times K}$, the rank-$R$ decomposition of $\boldsymbol{\mathcal{X}}$ takes the form of a ternary product between three factor matrices:
\begin{equation}
\label{eq:tf1}
    \boldsymbol{\mathcal{X}} \approx [\![\mathbf{A}, \mathbf{B}, \mathbf{C}]\!] \equiv  \sum_{r=1}^R \mathbf{a}_{:r} \otimes \mathbf{b}_{:r} \otimes \mathbf{c}_{:r}
\end{equation}
where $\mathbf{a}_{:r} \in \mathbb{R}^I$, $\mathbf{b}_{:r} \in \mathbb{R}^J$ and $\mathbf{c}_{:r} \in \mathbb{R}^K$ are the columns of the latent factor matrices $\mathbf{A} \in \mathbb{R}^{I \times R}$, $\mathbf{B} \in \mathbb{R}^{J \times R}$ and $\mathbf{C} \in \mathbb{R}^{K \times R}$ and $\otimes$ denotes the outer product.
When $R$ is the rank of $\mathcal{X}$, Eq.~(\ref{eq:tf1}) holds with an equality, and the above operation is called Canonical Polyadic (CP) decomposition.
Elementwise the previous relation is written as:
\begin{equation}
\label{eq:tf2}
    (\boldsymbol{\mathcal{X}})_{ijk} \approx [\![\mathbf{a}_i, \mathbf{b}_j, \mathbf{c}_k]\!] \equiv  \sum_{r=1}^R \mathbf{A}_{ir}\mathbf{B}_{jr}\mathbf{C}_{kr}
\end{equation}
where $\mathbf{a}_i$, $\mathbf{b}_j$, $\mathbf{c}_k \in \mathbb{R}^R$ are rows of the factor matrices. For 2nd-order tensors (matrices) the operation is equivalent to the low-rank matrix decomposition ($\mathbf{X} \approx \mathbf{A}\mathbf{B}^\mathrm{T}$).\\
For a generic $N$-order tensor $\boldsymbol{\mathcal{X}} \in \mathbb{R}^{I_1 \times I_2 \times \dots \times I_N}$, low-rank decomposition is expressed as:
\begin{equation}\label{eq:multi-product}
    (\boldsymbol{\mathcal{X}})_{i_1i_2\dots i_N} \approx [\![\mathbf{a}^{(1)}_{i_1}, \mathbf{a}^{(2)}_{i_2},\dots, \mathbf{a}^{(N)}_{i_N}]\!] \equiv \sum_{r=1}^R \mathbf{A}^{(1)}_{i_1r}~\mathbf{A}^{(2)}_{i_2r}\dots \mathbf{A}^{(N)}_{i_Nr}
\end{equation}
where $\mathbf{a}^{(1)}_{i_1},\mathbf{a}^{(2)}_{i_2},\dots,\mathbf{a}^{(N)}_{i_N} \in \mathbb{R}^R$ ($i_n \in \{1,\dots,I_n\},~n \in \{1,\dots,N\}$) are rows of factor matrices $\mathbf{A}^{(1)} \in \mathbb{R}^{I_1 \times R}$, $\mathbf{A}^{(2)} \in \mathbb{R}^{I_2 \times R}$, $\dots, \mathbf{A}^{(N)} \in \mathbb{R}^{I_N \times R}$.

\subsection{Skip-gram with negative sampling (SGNS)}
The skip-gram approach was initially proposed in \wordvec~\cite{mikolov2013distributed} to obtain low-dimensional representations of words.
Starting from a textual corpus of words $w_1,w_2,\dots, w_m$ from a vocabulary $\mathcal{V}$,
it assigns to each word $w_s$ a context corresponding to words $w_{s-T},\dots,w_{s-1},w_{s+1},\dots,w_{s+T}$ surrounding $w_s$ in a window of size $T$.
Then a set of training samples $\mathcal{D} = \{(i,j),~i \in \mathcal{W},~j \in \mathcal{C}\}$ is built by collecting all the observed word-context pairs,  where $\mathcal{W}$ and $\mathcal{C}$ are the vocabularies of words and contexts respectively (normally $\mathcal{W} = \mathcal{C} = \mathcal{V}$).
Here we denote as $\#(i,j)$ the number of times $(i,j)$ appears in $\mathcal{D}$. Similarly we use $\#i = \sum_{j}\#(i,j)$ and
$\#j = \sum_{i}\#(i,j)$ as the number of times each word occurs in $\mathcal{D}$,
with relative frequencies  $P_{\mathcal{D}}(i,j)= \frac{\#(i,j)}{|\mathcal{D}|}$,
$P_{\mathcal{D}}(i)= \frac{\#i}{|\mathcal{D}|}$ and
$P_{\mathcal{D}}(j)= \frac{\#j}{|\mathcal{D}|}$.\\
SGNS computes $d$-dimensional representations for words and contexts in two matrices $\mathbf{W} \in \mathbb{R}^{|\mathcal{W}| \times d}$ and $\mathbf{C} \in \mathbb{R}^{|\mathcal{C}| \times d}$, performing a binary classification task in which pairs $(i,j) \in \mathcal{D}$ are positive examples and pairs $(i,j_\mathcal{N})$ with randomly sampled contexts are negative examples.
The probability of the positive class is
parametrized as the sigmoid ($\sigma(x) =  (1+e^{-x})^{-1}$) of the inner product of embedding vectors:
\begin{equation}
\label{eq:sgns1}
    P[~(i,j) \in \mathcal{D} ~|~\mathbf{w}_i,\mathbf{c}_j~] = \sigma(\mathbf{w}_i\cdot\mathbf{c}_j) = \sigma \Big(~ (\mathbf{W}\mathbf{C}^{\mathrm{T}})_{ij}\Big)
\end{equation}
and each word-context pair $(i,j)$ contributes to the loss as follows:
\begin{align}
\label{eq:sgns2}
    \ell(i,j) = & \log \sigma(\mathbf{w}_i\cdot\mathbf{c}_j) + \sum_{j_\mathcal{N} \sim P_\mathcal{N}}^\kappa \log [1-\sigma(\mathbf{w}_i\cdot\mathbf{c}_{j_\mathcal{N}})]\\
    \simeq & \log \sigma(\mathbf{w}_i\cdot\mathbf{c}_j) + \kappa\cdot \underset{j_\mathcal{N} \sim P_\mathcal{N}}{\mathbb{E}}[\log \sigma(-\mathbf{w}_i\cdot\mathbf{c}_{j_\mathcal{N}})]
\end{align}
where the second expression uses the symmetry property $\sigma(-x) = 1-\sigma(x)$ inside the expected value and $\kappa$ is the number of negative examples, sampled according to the empirical distribution of contexts $P_\mathcal{N}(j) = P_{\mathcal{D}}(j)$. In the original formulation of \wordvec,~negative samples are picked from a smoothed distribution  $P_\mathcal{N}(j) = \frac{(\#j)^{3/4}}{\sum_{j'}(\#j')^{3/4}}$ instead of the unigram probability $\frac{\#j}{|\mathcal{D}|}$,  but this smoothing has not been proved to have positive effects in graph representations.
\\
Following results found in \cite{levy2014neural},
the sum of all $\ell(i,j)$ weighted with the probability each pair $(i,j)$ appears in $\mathcal{D}$ gives the objective function asymptotically optimized:
\begin{align}\label{eq:sgns3}
    \mathcal{L}^{SGNS} = & - \sum_{i=1}^{ |\mathcal{W}|}\sum_{j=1}^{|\mathcal{C}|}P_{\mathcal{D}}(i, j)\Big[ \log \sigma(\mathbf{w}_i\cdot\mathbf{c}_j) + \kappa\cdot \underset{j_\mathcal{N} \sim P_\mathcal{N}}{\mathbb{E}}[\log \sigma(-\mathbf{w}_i\cdot\mathbf{c}_{j_\mathcal{N}}) \Big]\\
   \dots = & - \sum_{i=1}^{ |\mathcal{W}|}\sum_{j=1}^{|\mathcal{C}|}\Big[P_{\mathcal{D}}(i, j) \log \sigma(\mathbf{w}_i\cdot\mathbf{c}_j)
    + \kappa\ P_\mathcal{N}(i, j) \log \sigma(-\mathbf{w}_i\cdot\mathbf{c}_j)\Big]
\end{align}
where
$P_{\mathcal{N}}(i, j) = P_{\mathcal{D}}(i) \cdot P_{\mathcal{D}}(j)$ is the probability of $(i,j)$ under assumption of statistical independence.
\\
In \cite{levy2014neural} it has been shown that SGNS local loss $\mathcal{L}(i,j)$
exhibits a global optimum with respect to the parameters $\mathbf{w}_i,\mathbf{c}_j$ that satisfies these relations:
\begin{equation}\label{eq:sgns4}
    \frac{\partial \mathcal{L}(i,j)}{\partial( \mathbf{w}_i\cdot\mathbf{c}_j)} = 0 ~~~\Leftrightarrow~~~
    (\mathbf{W}\mathbf{C}^{\mathrm{T}})_{ij} \approx \log \left(\frac{P_{\mathcal{D}}(i, j)}{\kappa~P_{\mathcal{N}}(i, j)}\right) =
    \mathrm{PMI}(i,j) - \log(\kappa)
\end{equation}
which tell us that SGNS optimization is equivalent to a rank-$d$ matrix decomposition of the word-context pointwise mutual information (PMI) matrix shifted by a constant. Such factorization is an approximation of the empirical PMI matrix since in the typical case $d \ll \mathrm{min}(|\mathcal{W}|,|\mathcal{C}|)$.

\subsection{Generalization of SGNS to higher-order representations}
SGNS can be generalized to learn $d$-dimensional embeddings from collections of higher-order co-occurrences.
Starting with $N$ vocabularies  $\big[\mathcal{V}_1, \mathcal{V}_2, \dots, \mathcal{V}_N \big]$ and a set of $N$-order tuples $\mathcal{D} = \{(i_1,i_2,\dots,i_N),~ i_1 \in \mathcal{V}_1,~i_2 \in \mathcal{V}_2,~ \dots, ~i_N \in \mathcal{V}_N\}$, the objective is to learn $N$ factor matrices $\mathbf{A}^{(1)} \in \mathbb{R}^{|\mathcal{V}_1| \times d}, \dots, \mathbf{A}^{(N)} \in \mathbb{R}^{|\mathcal{V}_N| \times d}$ which summarize the co-occurrence statistics of $\mathcal{D}$.

Keeping an example $(i_1,i_2,\dots,i_N) \in \mathcal{D}$, we define the loss with negative sampling scheme fixing $i_1$ and picking negative tuples $(\nu_2, \dots, \nu_N)$ according to the noise distribution $P_\mathcal{N}(\nu_2, \dots, \nu_N) = \prod_{n=2}^N \frac{\#\nu_n}{|\mathcal{D}|}\equiv\prod_{n=2}^N P_{\mathcal{D}}(\nu_n)$:
\begin{align*}
\centering
\small
    \ell(i_1, i_2, \dots, i_N) =  \log \sigma \big([\![\mathbf{a}^{(1)}_{i_1},&\mathbf{a}^{(2)}_{i_2}, \dots,\mathbf{a}^{(N)}_{i_N}]\!]\big) ~+\\&+~ \kappa\cdot  \underset{\nu_2, \dots, \nu_N \sim P_\mathcal{N}}{\mathbb{E}}\Big[\log \sigma\big(-[\![\mathbf{a}^{(1)}_{i_1},\mathbf{a}^{(2)}_{\nu_2}, \dots,\mathbf{a}^{(N)}_{\nu_N}]\!]\big)\Big]
\end{align*}
where each embedding $\mathbf{a}_{i_n}^{(n)}$ is the $i_n$-th row of the matrix $\mathbf{A}^{(n)}$.
The expectation term can be explicited:
\begin{equation*}
\centering
\small
    \underset{\nu_2, \dots, \nu_N \sim P_\mathcal{N}}{\mathbb{E}}\Big[\log \sigma\big(-[\![\mathbf{a}^{(1)}_{i_1},\mathbf{a}^{(2)}_{\nu_2}, \dots,\mathbf{a}^{(N)}_{\nu_N}]\big)\Big] = \sum_{j_2,\dots,j_N} P_\mathcal{N}(j_2,\dots,j_N) \log \sigma\big(-[\![\mathbf{a}^{(1)}_{i_1},\mathbf{a}^{(2)}_{j_2}, \dots,\mathbf{a}^{(N)}_{j_N}]\!]\big)
\end{equation*}
Weighting the loss error for each tuple $(i_1,i_2,\dots,i_N)$ with their  empirical probability $P_\mathcal{D}(i_1,i_2,\dots,i_N) = \frac{\#(i_1, i_2, \dots, i_N)}{|\mathcal{D}|}$,
and defining $[\![\mathbf{a}^{(1)}_{i_1},\mathbf{a}^{(2)}_{i_2}, \dots,\mathbf{a}^{(N)}_{i_N}]\!] \equiv m_{{i_1}{i_2}\dots{i_N}}$, we obtain the global objective with the sum over all combinations of vocabulary elements:
\begin{equation*}
\centering
\small
\mathcal{L} = - \sum_{i_1,i_2,\dots,i_N} P_\mathcal{D}(i_1,i_2,\dots,i_N) \Big[\log \sigma (m_{{i_1}{i_2}\dots{i_N}}) + \kappa \sum_{j_2,\dots,j_N} P_\mathcal{N}(j_2,\dots,j_N) \log \sigma(-m_{{i_1}{j_2}\dots{j_N}})\Big]
\end{equation*}\useshortskip
\begin{align*}
\centering
\small
= - \sum_{i_1,i_2,\dots,i_N} P_\mathcal{D}(i_1,i_2&,\dots,i_N)\log \sigma (m_{{i_1}{i_2}\dots{i_N}}) ~+ \\
-~&\kappa \sum_{i_1,i_2,\dots,i_N} P_\mathcal{D}(i_1,i_2,\dots,i_N)\sum_{j_2,\dots,j_N} P_\mathcal{N}(j_2,\dots,j_N) \log \sigma (-m_{{i_1}{j_2}\dots{j_N}})
\end{align*}

In the second term we can notice that only $P_\mathcal{D}(i_1,i_2,\dots,i_N)$ depends on the $N-1$ indices $(i_2,\dots,i_N)$, so performing the sum over that subset of indices we obtain the marginal distribution $\sum_{i_2 \dots i_N} P_\mathcal{D}(i_1,i_2,\dots,i_N) = P_\mathcal{D}(i_1)$. Finally renaming indices $\{j_h\} \rightarrow \{i_h\}$ and observing that $P_\mathcal{D}(i_1)P_\mathcal{N}(i_2,\dots,i_N) \equiv P_\mathcal{N}(i_1,i_2,\dots,i_N)$, we obtain the final loss:
\begin{equation}\label{eq:hosgnsobj}
\centering
\small
\mathcal{L}^{HOSGNS} = - \sum_{i_1,\dots,i_N} \Big[P_\mathcal{D}(i_1,\dots,i_N) \log \sigma (m_{{i_1}\dots{i_N}}) + \kappa \cdot P_\mathcal{N}(i_1,\dots,i_N) \log \sigma (-m_{{i_1}\dots{i_N}})\Big]
\end{equation}
In particular for the 3rd-order and 4th-order cases, with vocabularies $\mathcal{V}_1=\mathcal{W},~ \mathcal{V}_2=\mathcal{C},~ \mathcal{V}_3=\mathcal{T},~ \mathcal{V}_4=\mathcal{S}$ and embedding matrices $\mathbf{A}^{(1)}=\mathbf{W},~ \mathbf{A}^{(2)}=\mathbf{C},~ \mathbf{A}^{(3)}=\mathbf{T},~ \mathbf{A}^{(4)}=\mathbf{S}$, we have the loss functions minimized by our time-varying graph embedding model:
\begin{equation*}
\centering
\small
    \mathcal{L}^{(3rd)} = - \sum_{i,j,k}\Big[ P_\mathcal{D}(i,j,k) \log \sigma \big([\![\mathbf{w}_i, \mathbf{c}_j, \mathbf{t}_k]\!]\big) + \kappa\ P_{\mathcal{N}}(i,j,k) \log \sigma \big(-[\![\mathbf{w}_i, \mathbf{c}_j, \mathbf{t}_k]\!]\big)\Big]
\end{equation*}
\begin{align*}
\centering
\small
    \mathcal{L}^{(4th)} = - \sum_{i,j,k,l}\Big[ P_\mathcal{D}(i,j,k,l) \log \sigma \big([\![\mathbf{w}_i,\mathbf{c}_j,& \mathbf{t}_k, \mathbf{s}_l]\!]\big) + \kappa\ P_{\mathcal{N}}(i,j,k,l) \log \sigma \big(-[\![\mathbf{w}_i, \mathbf{c}_j, \mathbf{t}_k, \mathbf{s}_l]\!]\big)\Big]
\end{align*}

\subsection{HOSGNS as implicit tensor factorization}

Here we show the equivalence of HOSGNS to low-rank tensor factorization of the shifted PMI tensor into factor matrices, which is a straightforward generalization of previous proofs done for SGNS.

\textbf{Theorem.} Let $\mathcal{D} = \{(i_1,i_2,\dots,i_N),~ i_1 \in \mathcal{V}_1,~i_2 \in \mathcal{V}_2,~ \dots, ~i_N \in \mathcal{V}_N\}$ a training set of higher-order co-occurrences and $\mathrm{PMI}(i_1,\dots,i_N) = \log \left( \frac{P_\mathcal{D}(i_1,\dots,i_N)}{P_\mathcal{N}(i_1,\dots,i_N)} \right)$ the entries of the pointwise mutual information tensor computed from $\mathcal{D}$. Let $\mathbf{A}^{(1)} \in \mathbb{R}^{|\mathcal{V}_1| \times d}, \dots, \mathbf{A}^{(N)} \in \mathbb{R}^{|\mathcal{V}_N| \times d}$ embedding matrices of $\mathrm{HOSGNS}$. For $d$ sufficiently large, $\mathrm{HOSGNS}$ has the same global optimum as the canonical polyadic decomposition of $\mathrm{SPMI}_{\kappa}$, the PMI tensor shifted by $\log \kappa$.
\begin{proof}
We consider each relation $[\![\mathbf{a}^{(1)}_{i_1}, \dots,\mathbf{a}^{(N)}_{i_N}]\!] \equiv m_{{i_1}\dots{i_n}}$ as a mapping from combinations of embedding vectors to elements of a tensor $\boldsymbol{\mathcal{M}} \in \mathbb{R}^{|\mathcal{V}_1| \times \dots \times |\mathcal{V}_N|}$. The global loss $\mathcal{L} = \sum_{{i_1}\dots{i_N}} \mathcal{L}(i_1,\dots,i_N)$ in Eq.~(\ref{eq:hosgnsobj}) is the sum of local losses computed from elements of $\boldsymbol{\mathcal{M}}$:
 \begin{equation*}
    \small
     \mathcal{L}(i_1,\dots,i_N) = -\big[ P_\mathcal{D}(i_1,\dots,i_N)\log \sigma (m_{{i_1}\dots{i_N}}) + \kappa~P_\mathcal{N}(i_1,\dots,i_N)\log \sigma (-m_{{i_1}\dots{i_N}})\big]\\
 \end{equation*}
For sufficiently large $d$ (i.e. allowing for a perfect reconstruction of $\mathrm{SPMI}_{\kappa}$), each $m_{{i_1}\dots{i_N}}$ can assume a value independently of the others, and
we can treat the loss function $\mathcal{L}$ as a sum of independent addends, restricting the optimization problem to looking at the local objective and its derivative respect to $m_{{i_1}\dots{i_N}}$:
\begin{align*}
    \small
    \frac{\partial\mathcal{L}(i_1,\dots,i_N)}{\partial m_{{i_1}\dots{i_N}}} &= \kappa~P_\mathcal{N}(i_1,\dots,i_N) \sigma (m_{{i_1}\dots{i_N}}) -P_\mathcal{D}(i_1,\dots,i_N) \big[1-\sigma (m_{{i_1}\dots{i_N}})\big]\\
    &= \big[P_\mathcal{D}(i_1,\dots,i_N) + \kappa~P_\mathcal{N}(i_1,\dots,i_N)\big]\sigma(m_{{i_1}\dots{i_N}}) - P_\mathcal{D}(i_1,\dots,i_N)
\end{align*}
where we have used $\frac{d \sigma}{d x} = \sigma(x) (1-\sigma(x))$. To compare the derivative with zero, we use the identities $P_\mathcal{D} = (P_\mathcal{D} + \kappa~P_\mathcal{N})(1+\frac{\kappa~P_\mathcal{N}}{P_\mathcal{D}})^{-1}$ and $(1+x)^{-1} = \sigma (\log x^{-1})$:
\begin{equation*}
\small
    \frac{\partial\mathcal{L}(i_1,\dots,i_N)}{\partial m_{{i_1}\dots{i_N}}} = \left[P_\mathcal{D}(i_1,\dots,i_N) + \kappa~P_\mathcal{N}(i_1,\dots,i_N)\right]\left[\sigma (m_{{i_1}\dots{i_N}})-\sigma\left(\log \frac{P_\mathcal{D}(i_1,\dots,i_N)}{\kappa~P_\mathcal{N}(i_1,\dots,i_N)}\right)\right]
\end{equation*}
from which it follows that the derivative is 0 when elements $m_{{i_1}\dots{i_N}}$ are equal to the shifted PMI tensor entries:
\begin{equation}\label{eq:hosgnscpd}
\small
\frac{\partial \mathcal{L}(i_1,\dots,i_N)}{\partial m_{{i_1}\dots{i_N}}} = 0 ~~~\Leftrightarrow~~~\sum_{r=1}^d \mathbf{A}^{(1)}_{i_1r}\dots \mathbf{A}^{(N)}_{i_Nr}  = \log\left( \frac{P_\mathcal{D}(i_1,\dots,i_N)}{\kappa~P_\mathcal{N}(i_1,\dots,i_N)}\right) = \mathrm{SPMI}_{\kappa}(i_1,\dots,i_N)
\end{equation}
Since we have assumed that $d$ is large enough to ensure an exact reconstruction of $\mathrm{SPMI}_{\kappa}$, and this is true if $d \approx R = \mathrm{rank}(\mathrm{SPMI}_{\kappa}$), Eq.~(\ref{eq:hosgnscpd}) is consistent with the canonical polyadic decomposition of the shifted PMI tensor.
\end{proof}

\textbf{Remark.} In the typical case when $d \ll R$, the tensor reconstruction of Eq.~(\ref{eq:hosgnscpd}) is not exact, since the tensor is compressed in lower-dimensional factor matrices, but it still holds as a low-rank approximation:
\begin{equation*}
\small
\sum_{r=1}^{d\ll R} \mathbf{A}^{(1)}_{i_1r}\dots \mathbf{A}^{(N)}_{i_Nr}  \approx \mathrm{SPMI}_{\kappa}(i_1,\dots,i_N)
\end{equation*}

\subsection{Parameter settings}
Unless otherwise declared, all the embeddings are trained with a dimension $d$ of $128$ for node classification and $192$ for temporal event reconstruction

\textbf{HOSGNS} variants were optimized with Adam \cite{kingma2014adam} fixing the negative samples weight $\kappa=5$, the sample size $B=50000$ and linearly decaying the learning rate from a starting value of $0.05$ for $10^4$ iterations. For $\boldsymbol{\mathcal{A}}^{(dyn)}$ we set the random walks context window $T=10$.

For \textbf{\dyane~} we optimized \nodevec~with default hyperparameters (which comprise the same value $\kappa$=5 for negative samples and the same context window size $T=10$ that we chose for HOSGNS). The number of SGD epochs is 1 since we did not observe any improvement in downstream tasks by increasing the number of epochs.

For \textbf{\dyngem}~the model is trained with SGD with momentum (learning rate $10^{-3}$ and momentum coefficient 0.99) for 100 iterations in the first time-step and 30 for the others. We set the internal layer sizes of the autoencoder to $[400, 250, d]$.

\textbf{\dtriad}~is trained with Adagrad (learning rate $10^{-1}$) with 100 epochs and negative/positive samples ratio set to 5. Coefficients $\beta_0$ and $\beta_1$ related to social homofily and temporal smoothness are seto to $0.1$.

Due to the stochastic nature of the training, each of the above embedding models is trained 5 times for more robust performance estimates in downstream tasks.

\subsection{Embedding space visualization}

One of the main advantages of HOSGNS is that it able to disentangle the role of nodes and time by learning representations of nodes and time intervals separately.
In this section, we include plots with two dimensional projections of these embeddings, made with UMAP~\cite{mcinnes2018umap} for manifold learning and non-linear dimensionality reduction.
With these plots, we show that the embedding matrices learned by HOSGNS$^{(stat)}$ and HOSGNS$^{(dyn)}$   approaches successfully capture both the structure and the dynamics of the time-varying graph.

\begin{table}[h]
\caption{Number of class components for each label in \LyonSchool~dataset.}
    \centering
    \begin{small}
    \begin{tabular}{lc}
         \toprule
        \multirow{2}{*}{Class name}& Number of children\\
        & or teachers\\
        \midrule
        CP-A & 23\\
        CP-B & 25\\
        CE1-A & 23\\
        CE1-B & 26\\
        CE2-A & 23\\
        CE2-B & 22\\
        CM1-A & 21\\
        CM1-B & 23\\
        CM2-A & 22\\
        CM2-B & 24\\
        Teachers & 10\\
        \bottomrule
    \end{tabular}
    \end{small}
    \label{tab:lyonschool}
\end{table}

\begin{figure}[p]
\makebox[\linewidth]{
\begin{subfigure}{.6\textwidth}
  \centering
  \includegraphics[width=0.8\linewidth]{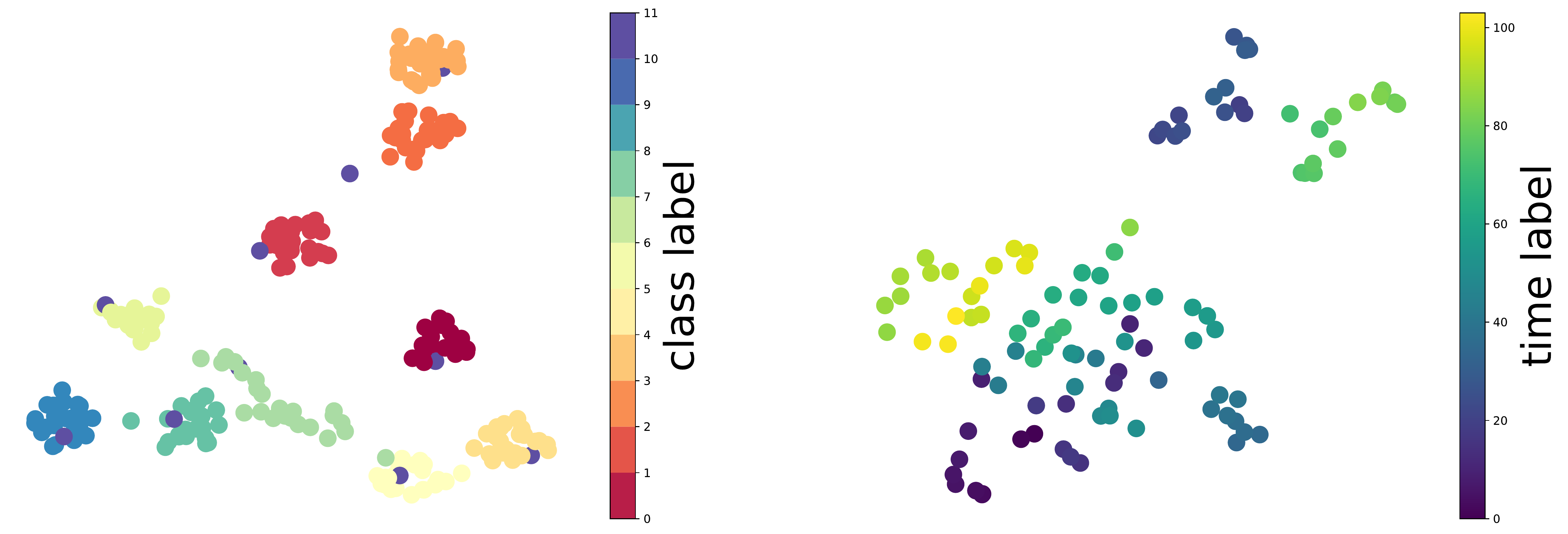}
  \caption{HOSGNS$^{(stat)}$}
\end{subfigure}
\begin{subfigure}{.6\textwidth}
  \centering
  \includegraphics[width=0.8\linewidth]{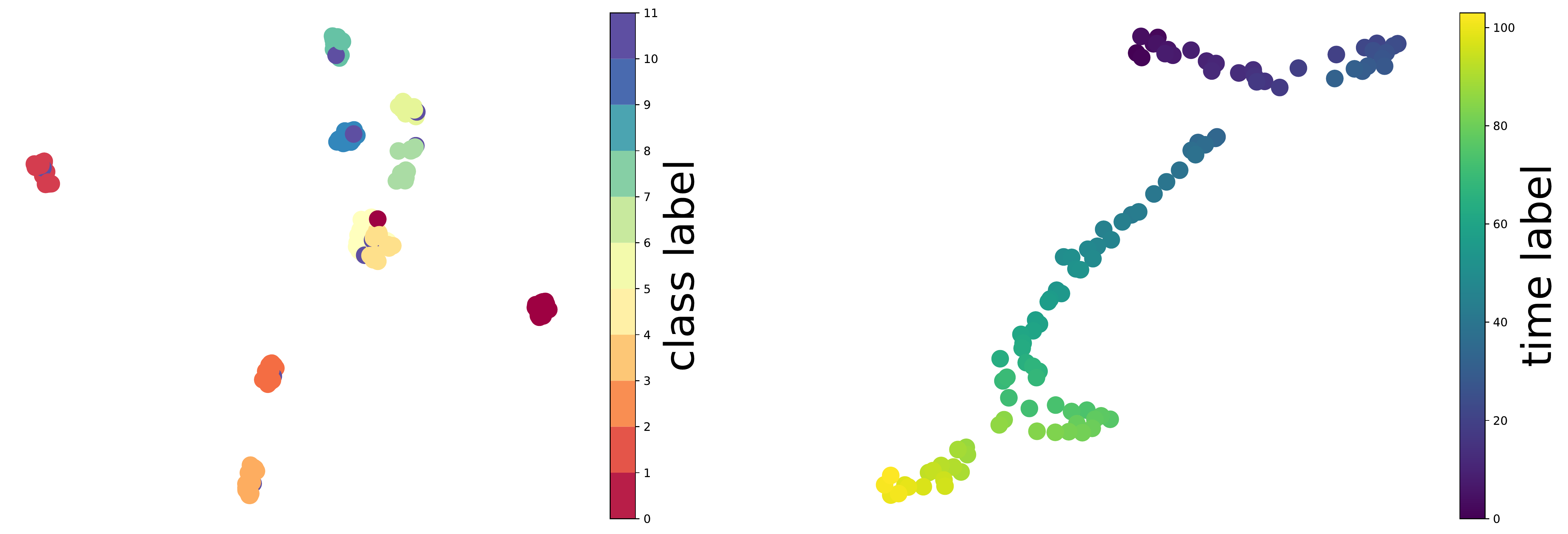}
  \caption{HOSGNS$^{(dyn)}$}
\end{subfigure}
}
\caption{Two-dimensional projection of the 128-dim embedding manifold spanned by embedding matrices $\mathbf{W}$ (left of each panel) and $\mathbf{T}$ (right of each panel), trained on \LyonSchool~data, of HOSGNS model trained on: (a) $\boldsymbol{\mathcal{P}}^{(stat)}$ and (b) $\boldsymbol{\mathcal{P}}^{(dyn)}$. These plots show how the community structure  and the evolution of time is captured by individual node embeddings $\{\mathbf{w}_i\}_{i \in \mathcal{V}}$  and time embeddings $\{\mathbf{t}_k\}_{k \in \mathcal{T}}$.}
\label{fig:embW_T}
\end{figure}

\begin{figure}[p]
\makebox[\linewidth]{
\begin{subfigure}{.6\textwidth}
  \centering
  \includegraphics[width=0.8\linewidth]{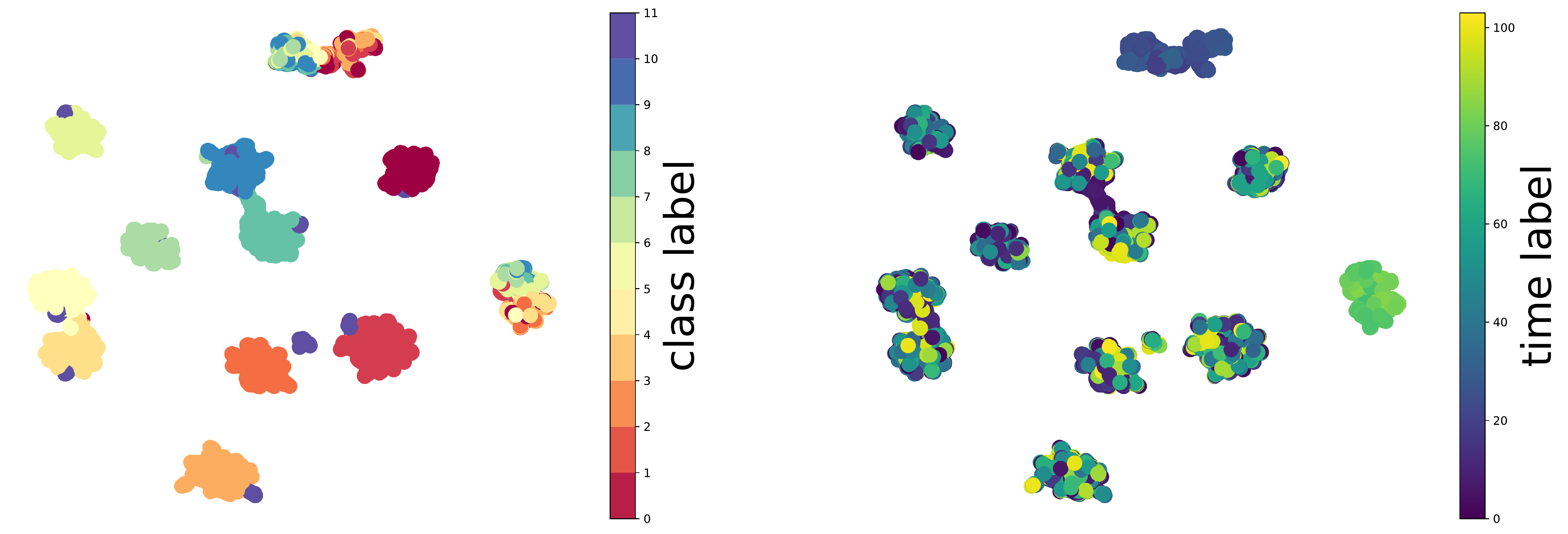}
  \caption{HOSGNS$^{(stat)}$}
\end{subfigure}%
\begin{subfigure}{.6\textwidth}
  \centering
  \includegraphics[width=0.8\linewidth]{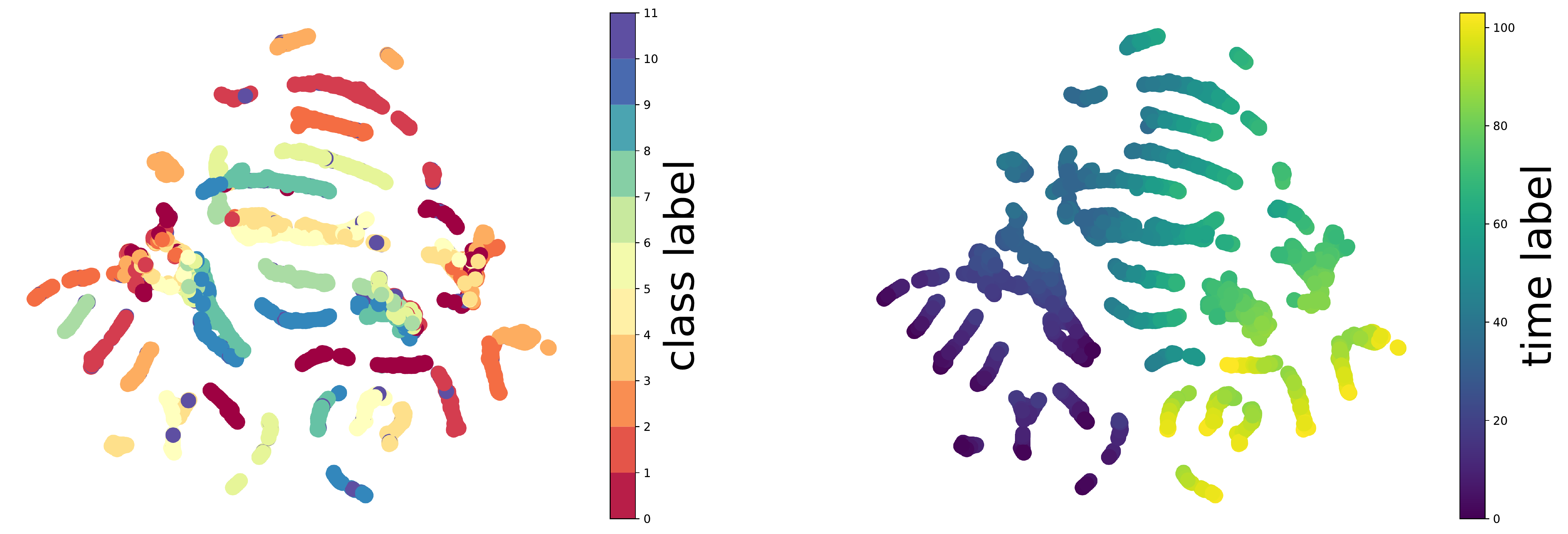}
  \caption{HOSGNS$^{(dyn)}$}
\end{subfigure}
}
\caption{Two-dimensional projection of the 128-dim embedding manifold spanned by dynamic node embeddings,  trained on \LyonSchool~data and obtained with Hadamard products $\{\mathbf{w}_i\circ\mathbf{t}_k\}_{(i,k) \in \mathcal{V}^{(\mathcal{T})}}$  between rows of $\mathbf{W}$ (node embeddings) and $\mathbf{T}$ (time embeddings), from HOSGNS model trained on: (a) $\boldsymbol{\mathcal{P}}^{(stat)}$ and (b) $\boldsymbol{\mathcal{P}}^{(dyn)}$. We highlight the temporal participation to communities (left of each panel) and the time interval of activation (right of each panel).}
\label{fig:embWT}
\end{figure}

\begin{figure}[p]
\makebox[\linewidth]{
\begin{subfigure}{.6\textwidth}
  \centering
  \includegraphics[width=0.8\linewidth]{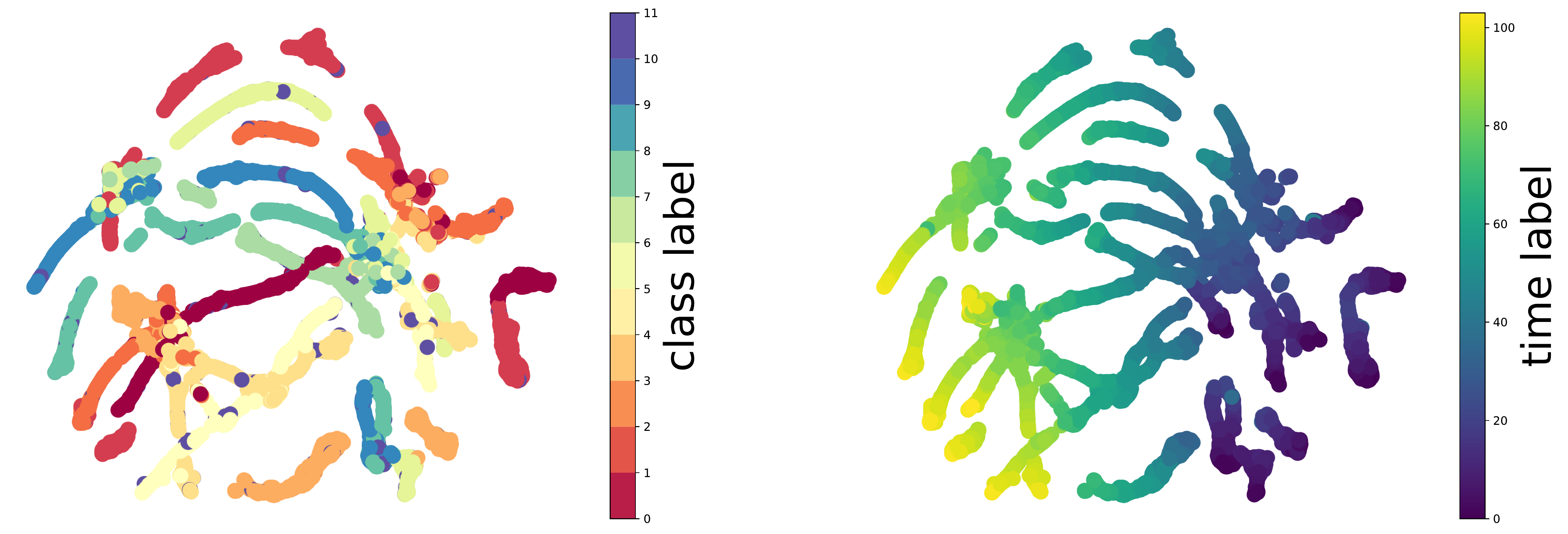}
  \caption{\dyane}
\end{subfigure}
\begin{subfigure}{.6\textwidth}
  \centering
  \includegraphics[width=0.8\linewidth]{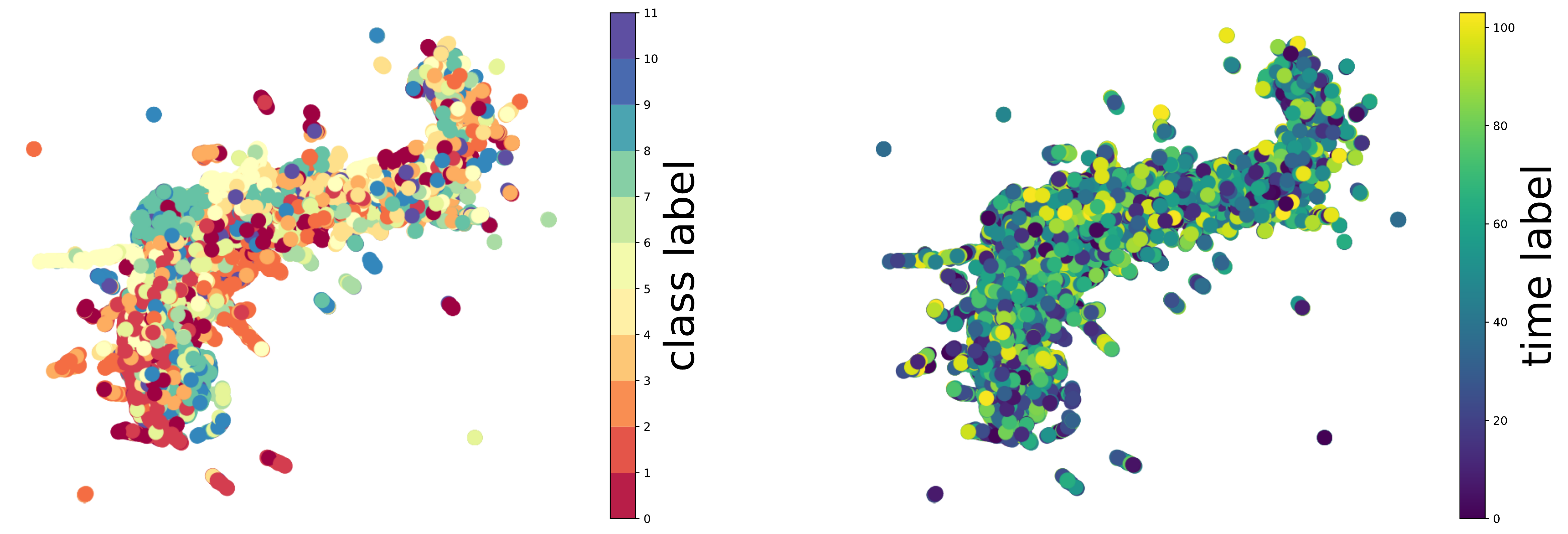}
  \caption{\dyngem}
\end{subfigure}
}
\centering
\begin{subfigure}{0.6\textwidth}
  \centering
  \includegraphics[width=0.8\linewidth]{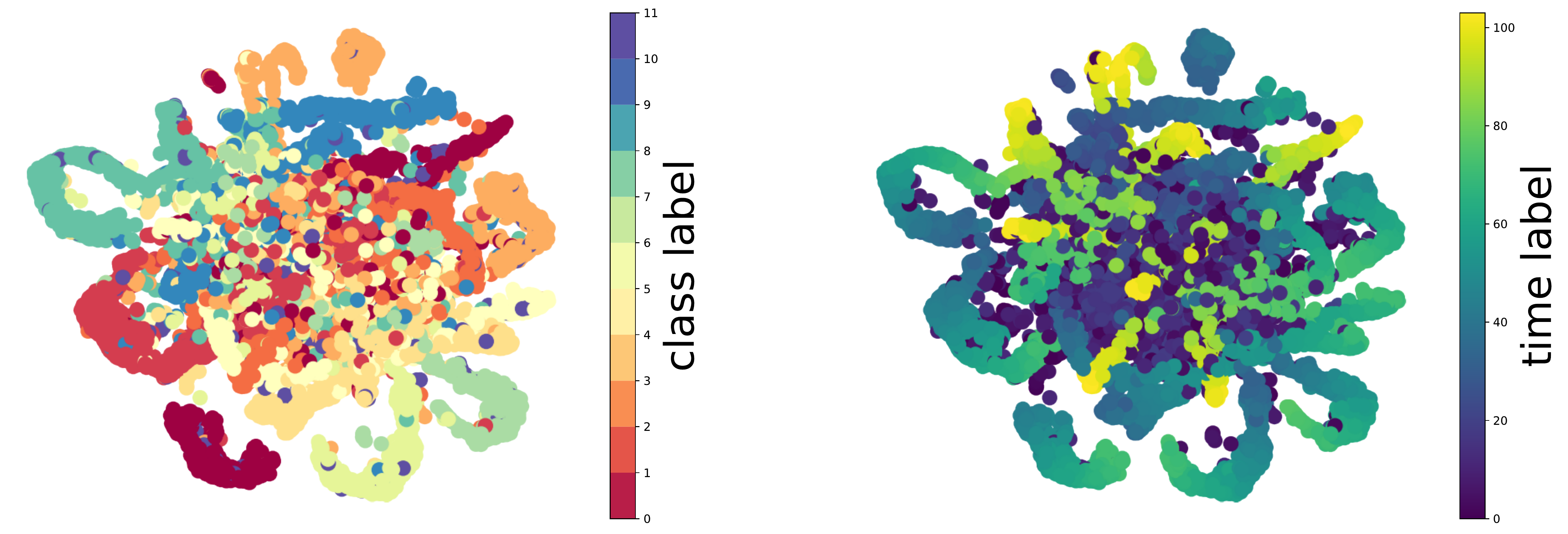}
  \caption{\dtriad}
\end{subfigure}
\caption{Two-dimensional projections of the 128-dim embedding manifold spanned by dynamic node embeddings for \LyonSchool~data learned with: (a) \dyane, (b) \dyngem~and (c) \dtriad. As in Figure~\ref{fig:embWT} we highlight the temporal participation to communities (left of each panel) and the time interval of activation (right of each panel).}
\label{fig:embWT_baseline}
\end{figure}

Temporal information can be represented by associating each embedding vector to its corresponding $k \in \mathcal{T}$, while graph structure can be represented by associating each embedding vector to a community membership. While community membership can be estimated by different community detection methods, we choose to use a dataset with ground truth data containing node membership information.\\
We consider in this section the \LyonSchool~dataset as a case study, widely investigated in literature respect to structural and spreading properties \cite{stehle2011high, barrat2013temporal, starnini2012random, panisson2013fingerprinting, sapienza2018estimating, galimberti2018mining}.
This dataset includes metadata (Table~\ref{tab:lyonschool}) concerning the class of each participant of the school (10 different labels for children and 1 label for teachers), and we identify the community membership of each individual according to these labels (\textit{class} labels). Moreover we also assign \textit{time} labels according to activation of individual nodes in temporal snapshots. To show how disentangled representations capture different aspects of the evolving graph, in Figure~\ref{fig:embW_T} we plot individual representations of nodes $i \in \mathcal{V}$ and time slices $k \in \mathcal{T}$ labeled according to the class membership and the time snapshot respectively. In Figure~\ref{fig:embWT} we visualize representations of temporal nodes $i^{(k)} \in \mathcal{V}^{(\mathcal{T})}$, computed as Hadamard products of nodes and time embeddings, in order to highlight both structural and dynamical aspects captured by the same set of embedding vectors. In Figure~\ref{fig:embWT_baseline} we see dynamic node embeddings computed with baseline methods without dissociating structure and time.

\subsection{Intrinsic and extrinsic evaluation of embedding representations}

\begin{figure}[p]
\begin{subfigure}{\textwidth}
  \centering
  \includegraphics[width=\linewidth]{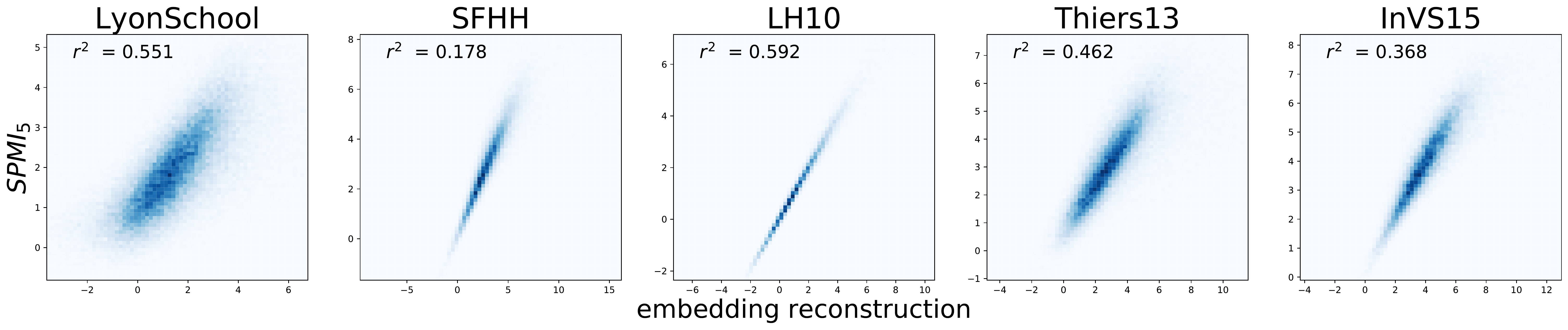}
  \caption{HOSGNS$^{(stat)}$}
\end{subfigure}
\begin{subfigure}{\textwidth}
  \centering
  \includegraphics[width=\linewidth]{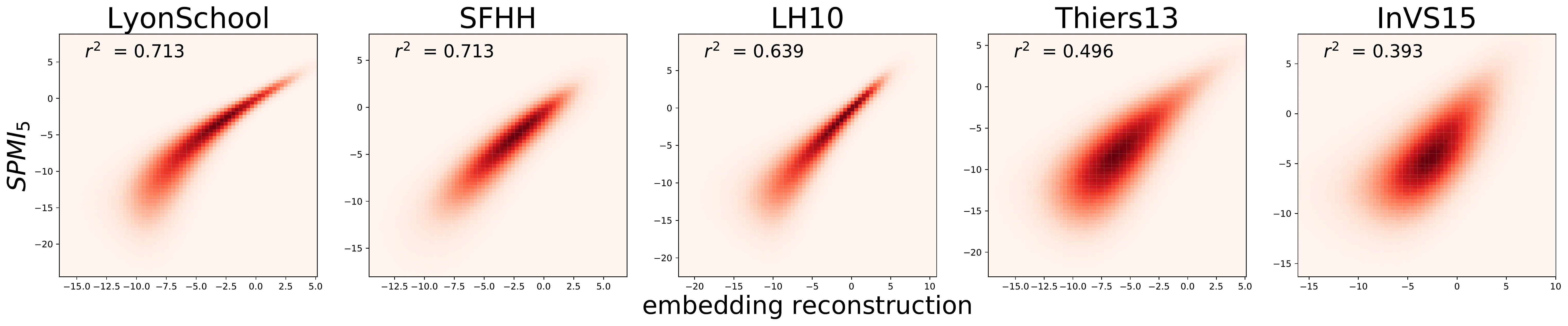}
  \caption{HOSGNS$^{(dyn)}$}
\end{subfigure}
\begin{subfigure}{\textwidth}
  \centering
  \includegraphics[width=\linewidth]{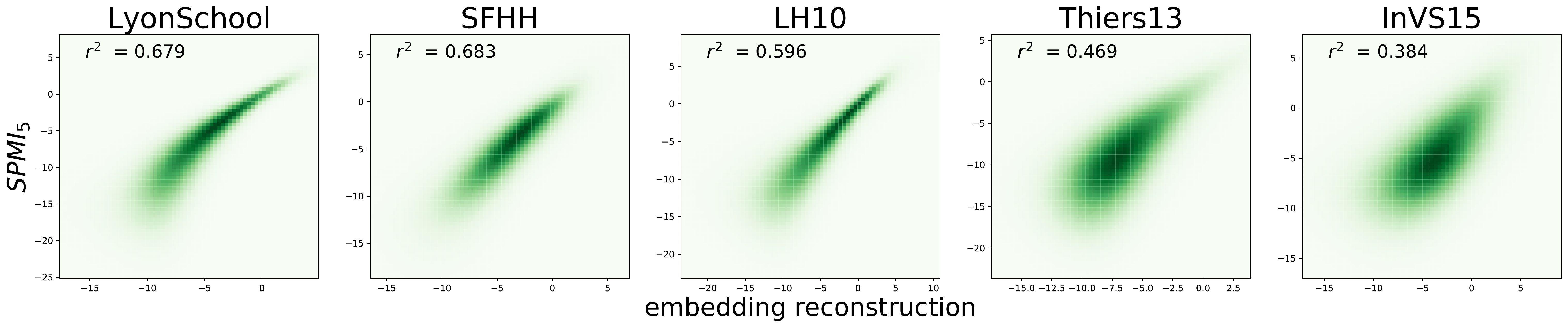}
  \caption{HOSGNS$^{(stat|dyn)}$}
\end{subfigure}
\caption{2D histograms of shifted PMI values $\mathrm{SPMI}_5(i,j,k\dots)$ (whereas are greater than $-\infty$) versus embedding reconstruction from higher-order inner products $[\![\mathbf{w}_i, \mathbf{c}_j, \mathbf{t}_k, \dots]\!]$, with HOSGNS models trained on: (a) $\boldsymbol{\mathcal{P}}^{(stat)}$, (b) $\boldsymbol{\mathcal{P}}^{(dyn)}$ and (c) $\boldsymbol{\mathcal{P}}^{(stat|dyn)}$.
The histograms were built by uniformly sampling $10^7$ entries from the $\mathrm{SPMI}_5$ tensors.
}
\label{fig:pmi}
\end{figure}

\begin{table}[!h]
    \caption{Operators and their definitions used to combine different embeddings learned with HOSGNS for tensors of order 3 (HOSGNS$^{(stat)}$) and 4 (HOSGNS$^{(dyn)}$ and HOSGNS$^{(stat|dyn)}$), applied to temporal node $i^{(k)}$ in node classification and to link $(i,j,k)$ in temporal event reconstruction. All operations, except \textit{Concat}, are described element-wise.}
    \label{tab:operators}
    \centering
    \begin{footnotesize}
    \makebox[\linewidth]{
    \begin{tabular}{llll}
    \toprule
    Operator & SGNS order &Node Classification& Temp. Event Reconstruction  \\
    \midrule
    Average & $3^{rd}, 4^{th}$& $\frac{1}{2}(\mathbf{w}_i+\mathbf{t}_k)$&
    $\frac{1}{3}(\mathbf{w}_i+\mathbf{c}_j+\mathbf{t}_k)$\\
    \midrule
    \multirow{2}{*}{Hadamard} & $3^{rd}$& $\mathbf{w}_i\circ\mathbf{c}_i\circ\mathbf{t}_k$&
    $\mathbf{w}_i\circ\mathbf{c}_j\circ\mathbf{t}_k$\\
    & $4^{th}$& $\mathbf{w}_i\circ\mathbf{t}_k$&
    $\mathbf{w}_i\circ\mathbf{c}_j\circ\mathbf{t}_k\circ\mathbf{s}_k$\\
    \midrule
    Weighted-L1 & $3^{rd}, 4^{th}$& $|\mathbf{w}_i-\mathbf{t}_k|$ &
    $\frac{1}{3}(|\mathbf{w}_i-\mathbf{t}_k|+|\mathbf{w}_i-\mathbf{c}_j|+|\mathbf{c}_j-\mathbf{t}_k|)$\\
    \midrule
    Weighted-L2 & $3^{rd}, 4^{th}$& $(\mathbf{w}_i-\mathbf{t}_k)^2$ &
    $\frac{1}{3}[(\mathbf{w}_i-\mathbf{t}_k)^2+(\mathbf{w}_i-\mathbf{c}_j)^2+(\mathbf{c}_j-\mathbf{t}_k)^2]$\\
    \midrule
    Concat & $3^{rd}, 4^{th}$& $[\mathbf{w}_i, \mathbf{t}_k]$ &
    $[\mathbf{w}_i, \mathbf{c}_j, \mathbf{t}_k]$\\
    \bottomrule
    \end{tabular}
    }
    \end{footnotesize}
\end{table}

\begin{figure}[p]
\begin{subfigure}{\textwidth}
  \centering
  \includegraphics[width=\linewidth]{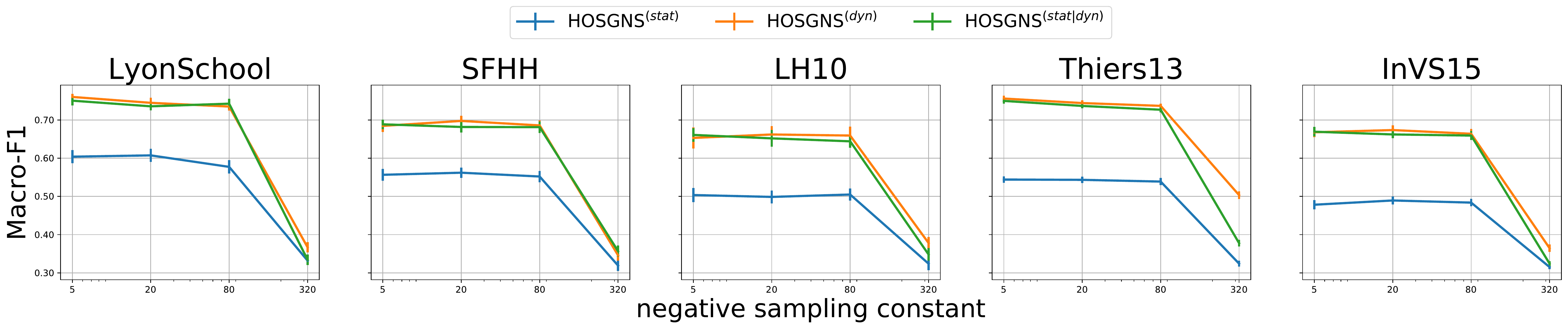}
  \caption{}
\end{subfigure}
\begin{subfigure}{\textwidth}
  \centering
  \includegraphics[width=\linewidth]{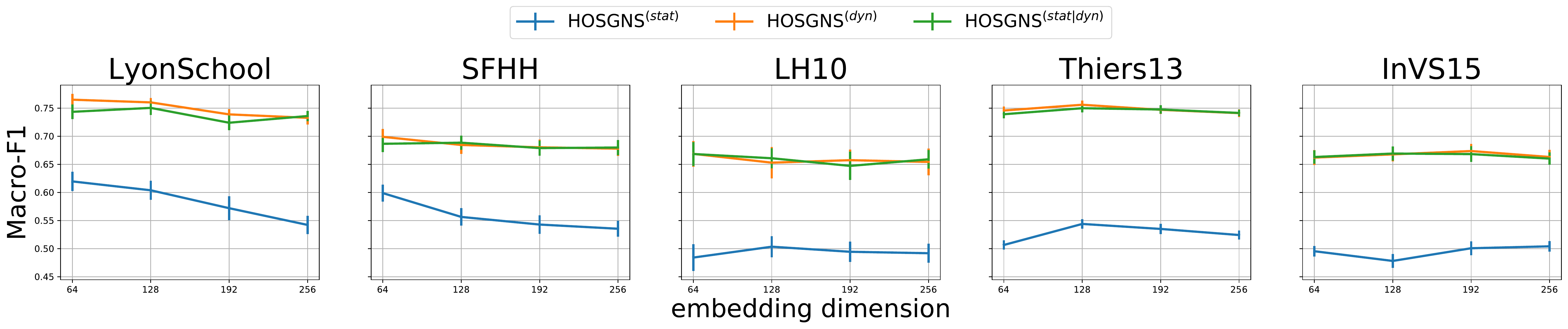}
  \caption{}
\end{subfigure}
\caption{Macro-F1 scores related to classification of nodes in SIR states from simulations with epidemic parameters $(\beta, \mu) = (0.125, 0.001)$, computed (a) fixing the embedding dimension to 128 and varying the negative sampling parameter $\kappa$ and (b) fixing $\kappa=5$ and varying the embedding dimension. In both panels time-resolved embedding vectors of nodes are computed with Hadamard product as explained in Table~\ref{tab:operators}.}
\label{fig:sir_analysis}
\end{figure}

\begin{figure}[p]
\begin{subfigure}{\textwidth}
  \centering
  \includegraphics[width=\linewidth]{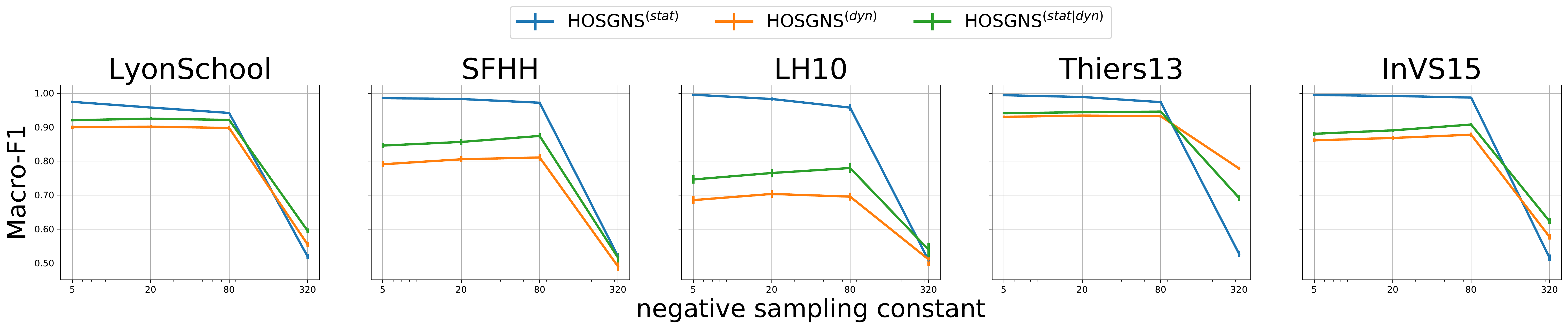}
  \caption{}
\end{subfigure}
\begin{subfigure}{\textwidth}
  \centering
  \includegraphics[width=\linewidth]{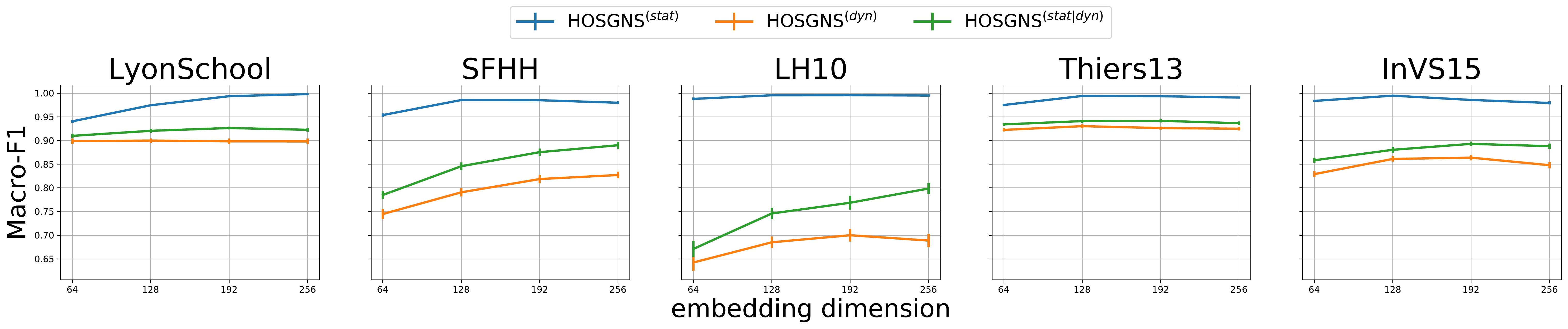}
  \caption{}
\end{subfigure}
\caption{Macro-F1 scores related to temporal event reconstruction, computed (a) fixing the embedding dimension to 128 and varying the negative sampling parameter $\kappa$ and (b) fixing $\kappa=5$ and varying the embedding dimension. In both panels time-resolved embedding vectors of edges are computed with Hadamard product as explained in Table~\ref{tab:operators}.}
\label{fig:link_analysis}
\end{figure}

Here we report results about intrinsic evaluations of the quality of embedding learned with HOSGNS, besides to completing the extrinsic evaluation in downstream tasks already reported (partially) in the main paper.

As intrinsic evaluation, in Figure~\ref{fig:pmi} we probe the capability of the model to reconstruct the shifted PMI tensor entries computing the higher order product of embedding vectors, operation optimized during the training phase to classify non-zero elements of the tensor itself. We verify the goodness of fit estimating the square of the Pearson coefficient between the distribution of actual PMI values and the estimated ones, having fixed the model $\kappa=5$ during training.

In Tables~\ref{tab:ncl}, \ref{tab:ncl_} and \ref{tab:lrec} we report Macro-F1 scores in downstream tasks, as extrinsic evaluation, with different operations used to construct embeddings for the logistic regression. For both node classification and temporal event reconstruction, in Table~\ref{tab:operators} we present definitions of different operators employed (Hadamard included, the only one displayed in the paper). For node classification, we show in Tables~\ref{tab:ncl} and \ref{tab:ncl_} results related to all tested combinations of epidemic parameters $(\beta, \mu)$ used to simulate SIR models.

In Figures~\ref{fig:sir_analysis} and \ref{fig:link_analysis} we report a sensitivity analysis with the effect of the embedding size $d$ and the negative sampling constant $\kappa$ on prediction performances in downstream tasks.
\begin{table}
\caption{Macro-F1 scores for classification of nodes in epidemic states according to a SIR model with parameters $(\beta,\mu)$ previously shown in the paper. Here for each HOSGNS variant we tested different operators to produce node-time representations, all with a dimension $d=128$, used as input to a Logistic Regression.
For each $(\beta,\mu)$ we highlight the two highest scores and underline the best one.}
\label{tab:ncl}
\centering
\makebox[\linewidth]{

\begin{tiny}
\begin{tabular}{cllccccc}
\toprule
\multirow{2}{*}{$(\beta,\mu)$}& \multirow{2}{*}{ Model}&\multirow{2}{*}{Operator}&\multicolumn{5}{c}{Dataset}\\
&&&\LyonSchool&SFHH&LH10&\Thiers&\InVS\\

\midrule\multirow{18}{*}{$(0.25,0.002)$}&\dyane&\multirow{3}{*}{-}&
          $77.8\pm1.4$&$66.7\pm2.0$&$54.7\pm2.4$&$\mathbf{73.2\pm1.2}$&$\mathbf{\underline{64.9}\pm1.1}$\\
&\dyngem&&$57.3\pm1.5$&$39.9\pm2.3$&$34.7\pm1.9$&$36.8\pm1.5$&$59.0\pm2.3$\\
&\dtriad&&$30.9\pm0.7$&$29.1\pm1.0$&$30.3\pm0.8$&$30.5\pm0.3$&$30.6\pm0.3$\\
\cline{2-8}

&\multirow{5}{*}{HOSGNS$^{(stat)}$}&Average&
            $54.3\pm2.0$&$50.6\pm1.5$&$50.7\pm2.5$&$54.6\pm1.6$&$51.8\pm1.5$\\
&&Hadamard&$60.1\pm2.1$&$55.8\pm1.5$&$50.0\pm2.1$&$49.9\pm1.8$&$46.4\pm1.0$\\
&&Weighted-L1&$51.5\pm2.2$&$45.1\pm1.3$&$49.3\pm1.8$&$45.4\pm1.2$&$44.5\pm1.3$\\
&&Weighted-L2&$53.7\pm2.1$&$44.5\pm1.7$&$47.8\pm2.0$&$45.5\pm1.3$&$45.4\pm1.5$\\
&&Concat&$70.8\pm1.6$&$62.9\pm2.2$&$55.6\pm2.1$&$61.9\pm2.0$&$56.2\pm1.3$\\
\cline{3-8}

&\multirow{5}{*}{HOSGNS$^{(dyn)}$}&Average&
                $71.4\pm1.3$&$65.3\pm1.9$&$\mathbf{65.9\pm2.5}$&$68.1\pm1.5$&$63.0\pm1.6$\\
&&Hadamard&$\mathbf{\underline{78.9}\pm1.1}$&$\mathbf{\underline{69.1}\pm1.4}$&$61.7\pm1.7$&$\mathbf{\underline{73.4}\pm1.2}$&$\mathbf{64.4\pm1.4}$\\
&&Weighted-L1&$74.7\pm1.6$&$66.2\pm1.8$&$61.3\pm2.3$&$71.9\pm1.1$&$62.4\pm1.4$\\
&&Weighted-L2&$74.3\pm1.5$&$66.4\pm1.5$&$61.2\pm2.4$&$70.6\pm1.3$&$62.6\pm1.3$\\
&&Concat&$76.3\pm1.1$&$68.2\pm1.6$&$\mathbf{\underline{68.5}\pm2.8}$&$71.6\pm1.1$&$64.7\pm1.5$\\
\cline{3-8}

&\multirow{5}{*}{HOSGNS$^{(stat|dyn)}$}
&Average&$72.1\pm1.2$&$64.4\pm1.3$&$63.0\pm2.2$&$69.0\pm1.4$&$61.3\pm1.3$\\
&&Hadamard&$\mathbf{78.6\pm1.1}$&$\mathbf{68.2\pm1.3}$&$61.6\pm2.3$&$72.2\pm1.3$&$63.8\pm1.4$\\
&&Weighted-L1&$73.9\pm1.1$&$64.6\pm1.4$&$60.6\pm2.2$&$70.8\pm1.3$&$62.8\pm1.9$\\
&&Weighted-L2&$71.1\pm2.3$&$64.1\pm1.6$&$60.0\pm1.9$&$70.7\pm1.5$&$61.4\pm1.8$\\
&&Concat&$75.6\pm1.3$&$67.4\pm1.9$&$64.9\pm2.5$&$70.5\pm1.1$&$64.0\pm1.8$\\

\midrule\multirow{18}{*}{$(0.125,0.001)$}&\dyane&\multirow{3}{*}{-}&$74.4\pm1.2$&$68.3\pm1.3$&$64.3\pm1.8$&$72.5\pm0.6$&$65.9\pm1.3$\\
&\dyngem&&$56.8\pm1.7$&$30.6\pm2.0$&$39.6\pm1.8$&$33.9\pm0.9$&$59.3\pm1.5$\\
&\dtriad&&$32.8\pm1.1$&$31.6\pm1.2$&$30.4\pm0.9$&$27.6\pm0.8$&$29.6\pm0.2$\\
\cline{2-8}

&\multirow{5}{*}{HOSGNS$^{(stat)}$}&Average&
            $55.0\pm1.5$&$53.1\pm1.5$&$49.8\pm2.6$&$59.4\pm1.0$&$55.0\pm1.3$\\
&&Hadamard&$60.4\pm1.7$&$55.7\pm1.6$&$50.4\pm1.9$&$54.4\pm0.9$&$47.8\pm1.2$\\
&&Weighted-L1&$50.9\pm1.6$&$46.5\pm1.1$&$51.4\pm1.8$&$48.5\pm0.9$&$45.1\pm1.2$\\
&&Weighted-L2&$52.8\pm1.7$&$46.6\pm1.1$&$48.3\pm1.8$&$48.0\pm1.1$&$44.6\pm1.1$\\
&&Concat&$66.7\pm1.6$&$61.0\pm1.9$&$55.0\pm2.6$&$65.4\pm1.1$&$59.1\pm1.0$\\
\cline{3-8}

&\multirow{5}{*}{HOSGNS$^{(dyn)}$}&Average&
            $69.3\pm1.1$&$65.3\pm1.3$&$\mathbf{66.5\pm2.0}$&$72.5\pm0.8$&$65.4\pm1.2$\\
&&Hadamard&$\mathbf{\underline{76.0}\pm0.8}$&$\mathbf{68.5\pm1.6}$&$65.3\pm2.8$&$\mathbf{\underline{75.6}\pm0.7}$&$\mathbf{66.8\pm1.3}$\\
&&Weighted-L1&$70.9\pm1.3$&$66.5\pm1.7$&$64.9\pm2.3$&$73.9\pm0.7$&$64.3\pm1.0$\\
&&Weighted-L2&$70.7\pm1.0$&$65.0\pm1.4$&$66.0\pm2.2$&$73.5\pm0.8$&$64.6\pm1.2$\\
&&Concat&$73.6\pm1.0$&$68.2\pm1.6$&$\mathbf{\underline{66.8}\pm2.5}$&$73.4\pm0.9$&$66.5\pm1.2$\\
\cline{3-8}

&\multirow{5}{*}{HOSGNS$^{(stat|dyn)}$}
&Average&$68.2\pm1.4$&$65.4\pm1.5$&$64.3\pm2.0$&$72.2\pm0.8$&$64.8\pm1.2$\\
&&Hadamard&$\mathbf{75.1\pm1.3}$&$\mathbf{\underline{68.9}\pm1.3}$&$66.1\pm1.8$&$\mathbf{75.0\pm0.7}$&$\mathbf{\underline{66.9}\pm1.2}$\\
&&Weighted-L1&$70.4\pm1.4$&$64.7\pm1.3$&$63.6\pm2.5$&$73.2\pm0.9$&$65.8\pm1.3$\\
&&Weighted-L2&$70.6\pm1.0$&$66.2\pm1.1$&$64.4\pm2.0$&$73.0\pm0.7$&$65.2\pm0.9$\\
&&Concat&$72.1\pm1.2$&$66.4\pm1.5$&$64.1\pm2.8$&$72.1\pm0.8$&$66.0\pm1.2$\\

\midrule\multirow{18}{*}{$(0.0625,0.002)$}&\dyane&\multirow{3}{*}{-}&$73.0\pm1.0$&$64.0\pm1.2$&$53.0\pm2.2$&$66.5\pm0.8$&$59.8\pm0.9$\\
&\dyngem&&$54.3\pm1.9$&$32.0\pm1.4$&$33.0\pm1.5$&$33.7\pm0.9$&$53.8\pm1.1$\\
&\dtriad&&$29.4\pm0.9$&$30.1\pm1.1$&$30.4\pm0.9$&$27.2\pm0.6$&$28.7\pm0.5$\\
\cline{2-8}

&\multirow{5}{*}{HOSGNS$^{(stat)}$}&Average&
            $56.1\pm1.4$&$50.9\pm1.4$&$46.1\pm1.8$&$55.5\pm0.9$&$52.3\pm1.0$\\
&&Hadamard&$58.5\pm1.8$&$51.6\pm1.2$&$46.0\pm1.5$&$49.4\pm0.8$&$46.5\pm0.8$\\
&&Weighted-L1&$49.6\pm1.7$&$44.5\pm0.9$&$46.7\pm1.7$&$45.5\pm0.9$&$43.9\pm0.8$\\
&&Weighted-L2&$50.3\pm1.5$&$43.9\pm1.2$&$45.1\pm1.8$&$44.9\pm0.8$&$44.0\pm0.7$\\
&&Concat&$66.4\pm1.5$&$56.1\pm1.9$&$49.3\pm2.2$&$61.1\pm0.9$&$53.0\pm0.8$\\
\cline{3-8}

&\multirow{5}{*}{HOSGNS$^{(dyn)}$}&Average&
            $70.0\pm1.3$&$61.7\pm1.4$&$\mathbf{\underline{60.3}\pm2.1}$&$67.9\pm0.8$&$\mathbf{\underline{60.4}\pm0.8}$\\
&&Hadamard&$\mathbf{\underline{74.4}\pm1.0}$&$\mathbf{\underline{65.1}\pm1.2}$&$56.8\pm1.8$&$\mathbf{\underline{68.4}\pm0.7}$&$59.6\pm0.9$\\
&&Weighted-L1&$71.3\pm1.1$&$63.7\pm1.1$&$58.2\pm2.2$&$67.5\pm0.7$&$58.8\pm0.9$\\
&&Weighted-L2&$71.0\pm1.4$&$63.5\pm1.1$&$55.6\pm2.1$&$67.7\pm0.8$&$58.6\pm1.0$\\
&&Concat&$\mathbf{73.6\pm0.9}$&$64.5\pm1.2$&$57.2\pm2.4$&$\mathbf{68.1\pm0.7}$&$\mathbf{60.2\pm0.9}$\\
\cline{3-8}

&\multirow{5}{*}{HOSGNS$^{(stat|dyn)}$}
&Average&$68.9\pm1.3$&$61.4\pm1.2$&$56.7\pm1.7$&$67.7\pm0.9$&$59.7\pm0.9$\\
&&Hadamard&$73.1\pm1.2$&$\mathbf{64.6\pm1.3}$&$56.9\pm1.9$&$67.9\pm0.7$&$59.4\pm1.0$\\
&&Weighted-L1&$70.9\pm1.6$&$62.5\pm1.1$&$57.0\pm2.0$&$66.8\pm1.0$&$58.0\pm0.9$\\
&&Weighted-L2&$70.9\pm1.0$&$62.2\pm1.1$&$55.5\pm2.3$&$67.8\pm0.8$&$58.2\pm0.9$\\
&&Concat&$72.4\pm1.3$&$64.0\pm1.3$&$\mathbf{59.4\pm2.4}$&$67.2\pm0.8$&$59.4\pm1.2$\\

\bottomrule
\end{tabular}

\end{tiny}
}

\end{table}

\begin{table}
\caption{Macro-F1 scores for classification of nodes in epidemic states according to a SIR model with other combinations $(\beta,\mu)$ not shown in the paper. Here for each HOSGNS variant we tested different operators to produce node-time representations, all with a dimension $d=128$, used as input to a Logistic Regression.
For each $(\beta,\mu)$ we highlight the two highest scores and underline the best one. In the case $(\beta,\mu)=(0.125,0.004)$ results for datasets LH10 and \InVS~are discarded since the SIR simulation does not meet the condition $|I_{|\mathcal{T}|/2}|\geq1$, as explained in \dyane.}
\label{tab:ncl_}
\centering
\makebox[\linewidth]{

\begin{tiny}
\begin{tabular}{cllccccc}
\toprule
\multirow{2}{*}{$(\beta,\mu)$}& \multirow{2}{*}{ Model}&\multirow{2}{*}{Operator}&\multicolumn{5}{c}{Dataset}\\
&&&\LyonSchool&SFHH&LH10&\Thiers&\InVS\\

\midrule\multirow{18}{*}{$(0.125,0.002)$}&\dyane&\multirow{3}{*}{-}&$\mathbf{76.6\pm1.3}$&$\mathbf{71.3\pm1.3}$&$55.6\pm2.1$&$71.6\pm0.9$&$61.4\pm1.4$\\
&\dyngem&&$58.8\pm2.0$&$33.9\pm1.7$&$35.7\pm1.6$&$32.8\pm0.8$&$57.5\pm1.7$\\
&\dtriad&&$30.6\pm0.9$&$30.2\pm1.1$&$30.1\pm0.7$&$28.3\pm0.5$&$29.8\pm0.3$\\
\cline{2-8}

&\multirow{5}{*}{HOSGNS$^{(stat)}$}&Average&
            $54.4\pm1.7$&$54.8\pm1.3$&$48.0\pm1.8$&$57.5\pm1.1$&$51.9\pm1.3$\\
&&Hadamard&$60.1\pm1.9$&$57.1\pm1.9$&$48.8\pm1.6$&$51.6\pm1.0$&$45.5\pm0.8$\\
&&Weighted-L1&$50.7\pm1.5$&$45.8\pm1.4$&$47.3\pm1.8$&$47.2\pm0.8$&$43.9\pm1.0$\\
&&Weighted-L2&$51.6\pm1.6$&$44.5\pm1.3$&$46.0\pm1.7$&$47.4\pm0.8$&$42.8\pm1.1$\\
&&Concat&$70.5\pm2.3$&$62.5\pm1.6$&$52.1\pm2.6$&$63.8\pm1.0$&$55.0\pm1.2$\\
\cline{3-8}

&\multirow{5}{*}{HOSGNS$^{(dyn)}$}&Average&
        $71.9\pm1.2$&$67.8\pm1.3$&$\mathbf{62.6\pm2.3}$&$70.9\pm0.8$&$61.4\pm1.4$\\
&&Hadamard&$\mathbf{\underline{77.2}\pm1.1}$&$71.1\pm1.2$&$62.1\pm2.2$&$\mathbf{73.8\pm1.0}$&$61.7\pm1.3$\\
&&Weighted-L1&$74.8\pm1.2$&$68.7\pm1.3$&$60.3\pm2.1$&$72.4\pm1.0$&$60.7\pm1.2$\\
&&Weighted-L2&$74.4\pm1.0$&$68.8\pm1.2$&$59.3\pm1.6$&$72.9\pm0.8$&$\mathbf{61.7\pm1.2}$\\
&&Concat&$76.0\pm1.3$&$\mathbf{\underline{71.5}\pm1.3}$&$\mathbf{\underline{62.9}\pm2.7}$&$72.4\pm0.9$&$\mathbf{\underline{62.0}\pm1.4}$\\
\cline{3-8}

&\multirow{5}{*}{HOSGNS$^{(stat|dyn)}$}
&Average&$71.8\pm1.0$&$67.0\pm1.2$&$60.6\pm2.4$&$71.4\pm1.0$&$60.6\pm0.9$\\
&&Hadamard&$76.2\pm1.5$&$70.7\pm1.1$&$61.4\pm1.9$&$\mathbf{\underline{74.0}\pm0.9}$&$61.6\pm1.1$\\
&&Weighted-L1&$73.5\pm1.4$&$69.3\pm1.2$&$61.7\pm2.4$&$72.2\pm0.8$&$60.3\pm1.3$\\
&&Weighted-L2&$71.9\pm1.6$&$67.8\pm1.4$&$58.2\pm2.0$&$72.1\pm0.9$&$61.4\pm1.3$\\
&&Concat&$74.7\pm1.7$&$69.4\pm1.4$&$62.5\pm2.6$&$72.2\pm0.8$&$61.6\pm1.4$\\

\midrule\multirow{18}{*}{$(0.1875,0.001)$}&\dyane&\multirow{3}{*}{-}&$74.4\pm1.2$&$\mathbf{\underline{69.7}\pm1.6}$&$\mathbf{\underline{66.0}\pm2.4}$&$73.3\pm0.7$&$\mathbf{66.4\pm1.4}$\\
&\dyngem&&$58.1\pm1.1$&$32.7\pm2.2$&$42.1\pm2.6$&$34.7\pm1.7$&$60.9\pm1.9$\\
&\dtriad&&$32.6\pm1.2$&$31.5\pm1.0$&$31.2\pm1.1$&$28.3\pm0.4$&$30.5\pm0.3$\\
\cline{2-8}

&\multirow{5}{*}{HOSGNS$^{(stat)}$}&Average&
            $54.0\pm1.9$&$53.4\pm1.8$&$49.4\pm2.6$&$58.5\pm1.4$&$52.8\pm1.1$\\
&&Hadamard&$60.5\pm2.4$&$55.4\pm1.6$&$47.9\pm2.1$&$52.0\pm1.0$&$47.3\pm1.1$\\
&&Weighted-L1&$50.4\pm1.6$&$45.0\pm1.2$&$48.6\pm2.1$&$46.7\pm1.2$&$46.7\pm1.4$\\
&&Weighted-L2&$52.0\pm1.6$&$45.1\pm1.2$&$45.9\pm2.6$&$46.9\pm1.0$&$46.1\pm1.3$\\
&&Concat&$66.7\pm2.0$&$62.2\pm2.0$&$51.6\pm2.5$&$63.0\pm1.7$&$58.8\pm1.1$\\
\cline{3-8}

&\multirow{5}{*}{HOSGNS$^{(dyn)}$}&Average&
            $69.0\pm1.3$&$65.4\pm1.4$&$\mathbf{63.5\pm2.7}$&$71.3\pm0.8$&$64.8\pm1.5$\\
&&Hadamard&$\mathbf{\underline{75.6}\pm0.9}$&$\mathbf{68.7\pm1.7}$&$63.0\pm2.2$&$\mathbf{\underline{75.3}\pm0.9}$&$\mathbf{\underline{67.4}\pm1.4}$\\
&&Weighted-L1&$72.0\pm1.7$&$66.1\pm1.4$&$63.0\pm2.8$&$73.1\pm0.9$&$65.1\pm1.4$\\
&&Weighted-L2&$71.2\pm1.2$&$66.1\pm1.6$&$58.9\pm2.7$&$72.9\pm0.9$&$65.0\pm1.4$\\
&&Concat&$72.8\pm1.2$&$68.0\pm1.8$&$61.8\pm2.6$&$73.4\pm0.9$&$66.0\pm1.3$\\
\cline{3-8}

&\multirow{5}{*}{HOSGNS$^{(stat|dyn)}$}
&Average&$67.1\pm2.0$&$65.1\pm1.5$&$62.4\pm2.8$&$70.4\pm0.8$&$64.5\pm1.2$\\
&&Hadamard&$\mathbf{74.9\pm1.1}$&$68.2\pm1.5$&$62.0\pm2.0$&$\mathbf{74.2\pm1.0}$&$66.1\pm1.3$\\
&&Weighted-L1&$71.1\pm1.2$&$64.9\pm1.5$&$60.1\pm2.3$&$71.5\pm1.0$&$64.0\pm1.7$\\
&&Weighted-L2&$70.2\pm1.0$&$65.8\pm1.6$&$61.0\pm2.7$&$72.0\pm0.9$&$65.1\pm1.2$\\
&&Concat&$72.2\pm1.3$&$67.3\pm1.9$&$59.4\pm2.6$&$72.1\pm1.1$&$65.2\pm1.5$\\

\midrule\multirow{18}{*}{$(0.125,0.004)$}&\dyane&\multirow{3}{*}{-}&$76.1\pm1.1$&$61.9\pm1.5$&$\multirow{3}{*}{-}$&$\mathbf{\underline{69.4}\pm0.9}$&\multirow{3}{*}{-}\\
&\dyngem&&$59.0\pm1.9$&$31.4\pm1.1$&&$36.1\pm0.9$&\\
&\dtriad&&$31.3\pm0.8$&$29.3\pm0.8$&&$29.4\pm0.4$&\\
\cline{2-8}

&\multirow{5}{*}{HOSGNS$^{(stat)}$}&Average&
            $56.0\pm2.3$&$50.4\pm1.2$&\multirow{5}{*}{-}&$54.3\pm1.0$&\multirow{5}{*}{-}\\
&&Hadamard&$60.2\pm2.1$&$49.6\pm1.3$&&$49.2\pm0.7$&\\
&&Weighted-L1&$50.0\pm1.9$&$42.9\pm1.2$&&$45.5\pm0.8$&\\
&&Weighted-L2&$52.4\pm1.7$&$41.7\pm1.4$&&$45.0\pm0.9$&\\
&&Concat&$69.6\pm1.5$&$56.6\pm2.1$&&$60.1\pm1.3$&\\
\cline{3-8}

&\multirow{5}{*}{HOSGNS$^{(dyn)}$}&Average&
            $72.8\pm2.0$&$62.1\pm1.3$&\multirow{5}{*}{-}&$67.5\pm0.9$&\multirow{5}{*}{-}\\
&&Hadamard&$\mathbf{\underline{77.0}\pm1.6}$&$\mathbf{63.6\pm1.4}$&&$\mathbf{69.3\pm1.0}$&\\
&&Weighted-L1&$74.1\pm1.4$&$61.7\pm1.4$&&$67.4\pm1.0$&\\
&&Weighted-L2&$75.2\pm1.4$&$61.5\pm1.5$&&$67.3\pm1.0$&\\
&&Concat&$\mathbf{76.8\pm1.2}$&$\mathbf{\underline{64.5}\pm1.3}$&&$68.3\pm0.9$&\\
\cline{3-8}

&\multirow{5}{*}{HOSGNS$^{(stat|dyn)}$}
&Average&$72.6\pm1.6$&$60.4\pm1.5$&\multirow{5}{*}{-}&$67.0\pm1.1$&\multirow{5}{*}{-}\\
&&Hadamard&$75.3\pm1.6$&$62.9\pm1.6$&&$68.5\pm1.0$&\\
&&Weighted-L1&$72.8\pm1.7$&$60.6\pm1.7$&&$67.2\pm0.8$&\\
&&Weighted-L2&$71.0\pm2.5$&$60.1\pm1.4$&&$67.2\pm1.0$&\\
&&Concat&$75.8\pm1.6$&$62.6\pm1.3$&&$68.6\pm0.8$&\\
\bottomrule
\end{tabular}

\end{tiny}
}

\end{table}

\begin{table}
\caption{Macro-F1 scores for temporal event reconstruction.
Here for each HOSGNS variant we tested different operators to produce link-time representations, all with a dimension $d=192$, used as input to a Logistic Regression.
We highlight in bold the best two overall
scores for each dataset. For baseline models we underline their highest score.}
\label{tab:lrec}
\centering
\makebox[\linewidth]{
\begin{tiny}
\begin{tabular}{llccccc}\toprule
\multirow{2}{*}{ Model}&\multirow{2}{*}{Operator}&\multicolumn{5}{c}{Dataset}\\
&&\LyonSchool&SFHH&LH10&\Thiers&\InVS\\
\midrule

\multirow{5}{*}{\dyane}
&Average&$56.6\pm0.9$&$52.7\pm1.2$&$53.2\pm1.6$&$51.2\pm0.8$&$52.3\pm1.0$\\
&Hadamard&$89.5\pm0.6$&$\underline{86.5}\pm1.2$&$\underline{73.9}\pm1.5$&$94.4\pm0.3$&$93.7\pm0.4$\\
&Weighted-L1&$89.8\pm0.5$&$83.2\pm1.1$&$72.1\pm1.4$&$95.1\pm0.3$&$94.5\pm0.4$\\
&Weighted-L2&$\underline{90.5}\pm0.6$&$84.2\pm1.0$&$72.5\pm1.4$&$\mathbf{\underline{95.2}\pm0.2}$&$\mathbf{\underline{94.7}\pm0.4}$\\
&Concat&$65.8\pm1.0$&$53.3\pm1.0$&$55.8\pm1.2$&$57.4\pm1.3$&$50.8\pm1.0$\\
\cline{2-7}

\multirow{5}{*}{\dyngem}
&Average&$57.8\pm0.8$&$56.9\pm1.1$&$\underline{54.1}\pm1.8$&$40.1\pm0.6$&$43.2\pm1.4$\\
&Hadamard&$\underline{62.1}\pm0.9$&$54.4\pm1.4$&$52.0\pm2.2$&$39.7\pm1.0$&$44.5\pm1.3$\\
&Weighted-L1&$58.6\pm0.6$&$52.7\pm1.2$&$49.9\pm1.8$&$\underline{41.5}\pm0.5$&$\underline{45.9}\pm1.1$\\
&Weighted-L2&$54.3\pm0.8$&$47.0\pm1.4$&$46.5\pm1.9$&$39.5\pm0.5$&$42.6\pm1.5$\\
&Concat&$60.4\pm0.7$&$\underline{58.2}\pm0.9$&$48.2\pm1.8$&$36.9\pm0.5$&$45.2\pm1.1$\\
\cline{2-7}

\multirow{5}{*}{\dtriad}&Average&$51.4\pm0.6$&$57.0\pm0.9$&$58.4\pm1.4$&$57.7\pm0.5$&$55.1\pm0.7$\\
&Hadamard&$60.9\pm0.5$&$58.7\pm0.8$&$58.6\pm1.3$&$62.2\pm0.4$&$64.3\pm0.7$\\
&Weighted-L1&$\underline{78.7}\pm1.0$&$72.4\pm0.8$&$75.5\pm1.1$&$70.7\pm0.7$&$78.3\pm0.6$\\
&Weighted-L2&$77.1\pm0.9$&$\underline{72.9}\pm1.3$&$\mathbf{\underline{77.0}\pm1.1}$&$\underline{72.3}\pm0.6$&$\underline{78.7}\pm0.7$\\
&Concat&$52.5\pm0.6$&$53.3\pm0.9$&$56.3\pm1.1$&$55.1\pm0.5$&$52.9\pm0.8$\\
\midrule

\multirow{5}{*}{HOSGNS$^{(stat)}$}
&Average&$59.7\pm0.8$&$53.0\pm1.0$&$53.1\pm1.4$&$55.6\pm1.1$&$51.6\pm0.9$\\
&Hadamard&$\mathbf{99.4\pm0.1}$&$\mathbf{98.5\pm0.2}$&$\mathbf{99.6\pm0.2}$&$\mathbf{99.4\pm0.1}$&$\mathbf{98.6\pm0.3}$\\
&Weighted-L1&$72.1\pm0.8$&$60.4\pm1.0$&$56.6\pm1.6$&$68.1\pm1.0$&$60.7\pm1.0$\\
&Weighted-L2&$71.4\pm0.8$&$61.4\pm1.0$&$56.3\pm1.5$&$68.8\pm0.8$&$57.1\pm0.8$\\
&Concat&$63.3\pm1.1$&$54.8\pm1.0$&$52.4\pm2.1$&$58.5\pm1.3$&$51.7\pm1.1$\\
\cline{2-7}

\multirow{5}{*}{HOSGNS$^{(dyn)}$}
&Average&$62.5\pm1.1$&$53.2\pm1.2$&$52.4\pm1.8$&$56.3\pm1.2$&$51.1\pm1.1$\\
&Hadamard&$89.8\pm0.6$&$81.9\pm0.9$&$70.0\pm1.3$&$92.6\pm0.3$&$86.4\pm0.6$\\
&Weighted-L1&$81.0\pm0.8$&$63.9\pm1.0$&$56.5\pm1.5$&$83.0\pm0.6$&$63.7\pm1.0$\\
&Weighted-L2&$79.9\pm0.9$&$63.7\pm1.3$&$55.9\pm1.9$&$79.8\pm1.0$&$59.4\pm1.2$\\
&Concat&$64.0\pm1.0$&$54.4\pm1.4$&$50.6\pm2.5$&$58.5\pm1.6$&$50.6\pm1.3$\\
\cline{2-7}

\multirow{5}{*}{HOSGNS$^{(stat|dyn)}$}
&Average&$62.4\pm1.2$&$54.0\pm0.9$&$51.2\pm1.7$&$56.9\pm1.4$&$51.1\pm1.1$\\
&Hadamard&$\mathbf{92.6\pm0.3}$&$\mathbf{87.5\pm0.8}$&$76.9\pm1.5$&$94.2\pm0.3$&$89.3\pm0.5$\\
&Weighted-L1&$81.6\pm0.8$&$63.7\pm1.1$&$57.6\pm1.3$&$83.6\pm0.6$&$64.8\pm0.9$\\
&Weighted-L2&$80.6\pm0.8$&$63.6\pm1.0$&$55.9\pm1.4$&$81.6\pm0.9$&$61.3\pm1.0$\\
&Concat&$62.9\pm1.0$&$53.9\pm1.2$&$51.8\pm2.6$&$58.0\pm1.5$&$49.9\pm1.3$\\
\bottomrule
\end{tabular}

\end{tiny}
}
\end{table}

\end{document}